\documentclass[lettersize,journal]{IEEEtran}
\usepackage{amsmath,amsfonts}
\usepackage{amssymb} 
\usepackage{nccmath}

\usepackage{algorithmic}
\usepackage{algorithm}
\usepackage{array}
\usepackage[caption=false,font=normalsize,labelfont=sf,textfont=sf]{subfig}
\usepackage{textcomp}
\usepackage{stfloats}
\usepackage{url}
\usepackage{verbatim}
\usepackage{graphicx}
\usepackage{cite}
\usepackage{xcolor}

\begin{document}

\title{

Enhancing Safety in Mixed Traffic: Learning-Based Modeling and Efficient Control of Autonomous and Human-Driven Vehicles
}


\author{Jie Wang$^{1}$, Yash Vardhan Pant$^{1}$, Lei Zhao$^{2}$, Micha\l{} Antkiewicz$^{1}$, Krzysztof Czarnecki$^{1}$ 
\thanks{$^{1}$ Jie Wang, Micha\l{} Antkiewicz, Yash Vardhan Pant, and Krzysztof Czarnecki are with the Electrical and Computer Engineering Department, University of Waterloo, Waterloo, ON, Canada. E-mail: {\tt\small jwangjie@outlook.com; \{michal.antkiewicz, yash.pant, krzysztof.czarnecki\}@uwaterloo.ca}.}.
\thanks{$^{2}$ Lei Zhao is with the Computer and Software Engineering Department, Polytechnique Montréal, Montréal, QC, Canada. E-mail: {\tt\small lei.zhao@polymtl.ca}.}}



\maketitle

\begin{abstract}
With the increasing presence of autonomous vehicles (AVs) on public roads, developing robust control strategies to navigate the uncertainty of human-driven vehicles (HVs) is crucial. This paper introduces an advanced method for modeling HV behavior, combining a first-principles model with Gaussian process (GP) learning to enhance velocity prediction accuracy and provide a measurable uncertainty. We validated this innovative HV model using real-world data from field experiments and applied it to develop a GP-enhanced model predictive control (GP-MPC) strategy. This strategy aims to improve safety in mixed vehicle platoons by integrating uncertainty assessment into distance constraints. Comparative simulation studies with a conventional model predictive control (MPC) approach demonstrated that our GP-MPC strategy ensures more reliable safe distancing and fosters efficient vehicular dynamics, achieving notably higher speeds within the platoon. By incorporating a sparse GP technique in HV modeling and adopting a dynamic GP prediction within the MPC framework, we significantly reduced the computation time of GP-MPC, marking it only 4.6\% higher than that of the conventional MPC. This represents a substantial improvement, making the process about 100 times faster than our preliminary work without these approximations. Our findings underscore the effectiveness of learning-based HV modeling in enhancing both safety and operational efficiency in mixed-traffic environments, paving the way for more harmonious AV-HV interactions.
\end{abstract}

\begin{IEEEkeywords}
Human-driven vehicle modeling, modeling uncertainty, Gaussian process, mixed vehicle platoon, model predictive control.
\end{IEEEkeywords}

\section{Introduction}
\label{sec:intro}
\begin{figure}
    \centerline{\includegraphics[trim=0cm 0cm 0.0cm 2cm, width=0.96\columnwidth]{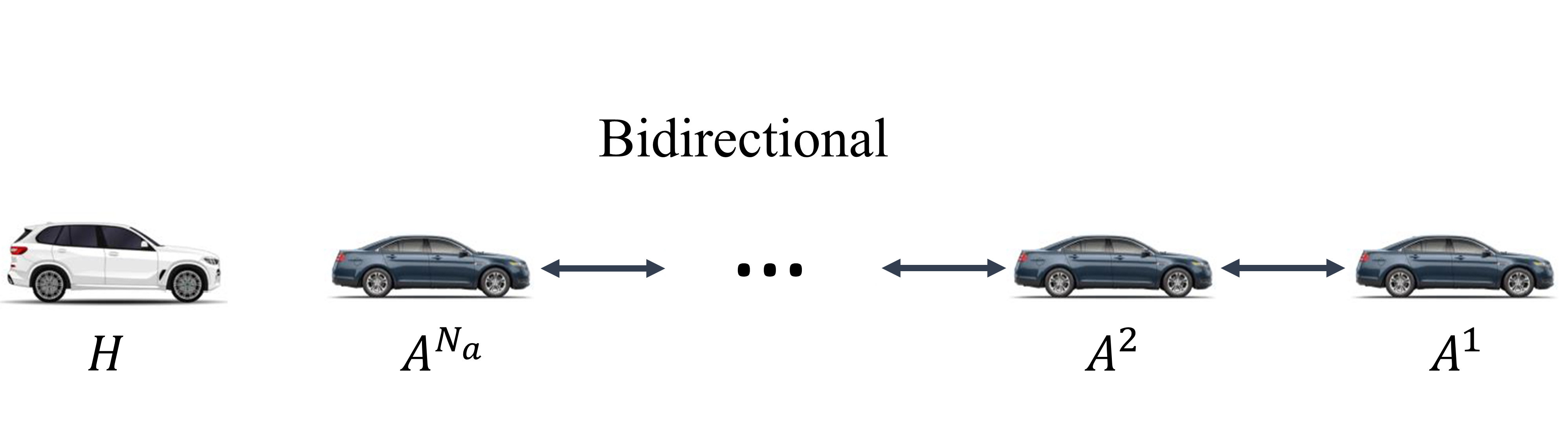}}
    \caption{A mixed vehicle platoon is composed of $N_a$ connected AVs, denoted as ${A^1, A^2, \cdots, A^{N_a}}$, trailed by a HV $H$. The AVs employ a sequential bidirectional communication topology for data sharing, but no direct communication takes place between the connected AVs and the HV. This configuration is motivated by recent studies highlighting that most accidents in mixed traffic involve HVs rear-ending AVs, leading to the specific platoon arrangement showcased.}
    \label{figure:mixed_platoon}
\end{figure}

\IEEEPARstart{T}{he} last ten years have seen remarkable advancements in the field of autonomous vehicles (AVs) and their integration into intelligent transport systems, including connected autonomous vehicle (CAV) platooning, with the primary goal of enhancing traffic safety and efficiency \cite{guo2023study}. CAV platooning is built upon the principle of synchronized driving, which allows for minimal space between vehicles while maintaining secure speeds. In such a setup, CAVs in a platoon can share real-time information such as speed and distance, thus facilitating synchronized control and coordinated movements. As a result, vehicles in a platoon can safely travel at higher speeds with significantly reduced gaps. Nonetheless, it is important to consider the continued prevalence of human-driven vehicles (HVs), which are expected to remain a significant part of the traffic ecosystem in the foreseeable future \cite{du2021cooperative,du2022framework}. This invariably leads to interactions between human and autonomous drivers, exemplified in Fig. \ref{figure:mixed_platoon}, where an HV trails a CAV platoon.

\subsection{Problem Statement and Objectives}
Mixed traffic environments, where AVs and HVs coexist, introduce significant challenges in vehicular control due to the uncertainty inherent in human driving behavior. Differences in reaction times, decision-making, and compliance with traffic laws between AVs and HVs amplify safety risks \cite{huang2020learning}. Analysis of recent traffic incidents in the United States has shown a marked increase in rear-end collisions involving HVs crashing into AVs in mixed traffic conditions, 64.2\% compared with 28.3\% in environments with only HVs \cite{petrovic2020}. These findings underscore an urgent need to bridge the behavioral understanding gap between AVs and HVs. Instead of relying on human drivers to drive cautiously around AVs, it would be more practical and effective to develop adaptive control strategies for CAVs that incorporate models of human driving behavior, thus enhancing safety in mixed-traffic environments.

In response to the notable issue of HVs rear-ending AVs, we aim to address this challenge by proposing a learning-based model to improve the prediction accuracy of human driving behaviors within mixed traffic scenarios. The primary objective is to leverage this model to develop an efficient control strategy for AVs that account for human uncertainties, enhancing safety and operational efficiency in mixed traffic conditions. Our focus is specifically on the dynamics of an HV following an AV platoon in longitudinal car-following scenarios, as illustrated in Fig. \ref{figure:mixed_platoon}.

\subsection{Contributions}
This paper presents a novel learning-based approach to model the behaviors of HVs in interactions with AVs by utilizing real-world data from field experiments. By incorporating a Gaussian process (GP) learning component into the HV modeling process, this research seeks to improve velocity prediction accuracy and introduce a quantifiable measure of uncertainty, thereby informing safer and more reliable longitudinal car-following control in mixed traffic scenarios. The key contributions of this research are summarized as follows:

\begin{itemize}
    \item We present a novel approach that integrates a first-principles model of human behaviors with a learning-based GP model trained on field experiment data to model HVs in a longitudinal velocity tracking setting. The GP model corrects the velocity predictions of the first-principles model, which was identified using human-in-the-loop simulator data. To ensure computational efficiency for subsequent control policy design, we employ a sparse GP technique to reduce the average computational time by 18 times compared to the standard GP model, while still resulting in a notable 36\% improvement in overall modeling accuracy over the first-principles model.

    \item We designed a chance-constrained model predictive control (GP-MPC) scheme to guarantee a safe distance in AV-HV interactions. This innovative GP-MPC utilizes the proposed HV model to account for modeling uncertainties via an additional probabilistic constraint. By blending this with a predefined deterministic distance, we formulate an adaptive safe distance constraint that enhances safety.

    \item Extensive simulation experiments demonstrate the superior performance of GP-MPC methodology over a baseline standard model predictive control (MPC) approach, which relies on the first-principles model. The GP-MPC not only assures a larger safety distance between vehicles but also improves the locomotion efficiency of the mixed platoon by enabling greater speeds for each vehicle. In real-time implementation settings, compared to the baseline MPC, the GP-MPC requires only a 4.6\% increase in computation time due to the employment of sparse GP modeling for HV and the dynamic GP prediction within the MPC strategy.   
\end{itemize}

\subsection{Related Work} 
\label{sec:related_work}

In the field of autonomous vehicle control, particularly in mixed-vehicle platoons involving both AVs and HVs \cite{xue2023,chen2021}, adaptive cruise control (ACC) and cooperative ACC have marked significant progress \cite{huang2020learning}. Car-following models play a crucial role in the development and implementation of these longitudinal control strategies. Among established models \cite{kessels2019traffic}, the Gazis-Herman-Rothery (GHR) model is one of the earliest yet foundational approaches. It defines a vehicle's acceleration based on its speed, the speed difference with the leading vehicle, and the headway (the distance to the leading vehicle), providing a simple but effective framework for basic automated following behaviors. Progressing toward more contemporary models, the intelligent driver model (IDM) offers a significant enhancement in modeling car-following behavior. It adapts a vehicle's acceleration by considering factors such as the current and desired speeds, minimum safe distance to the leading vehicle, and the relative speed. The optimal velocity model (OVM) focuses on determining an optimal velocity based on the distance to the preceding vehicle, thereby guiding the acceleration decisions in a way that is simple yet effective for maintaining stable traffic flow. The full velocity difference model (FVDM) expands OVM by incorporating a more comprehensive consideration of the velocity difference between the following and the leading vehicles. This enhanced approach is beneficial for vehicle platooning, where the behavior of multiple vehicles in a platoon needs to be harmonized for optimal flow and safety. These models focus on capturing the dynamics of a following vehicle based on the actions of a leading vehicle and emphasize reaction times, safe following distances, and relative velocities \cite{li2023survey}. 

However, when modeling HV-AV interaction in mixed traffic situations, these conventional models assume deterministic behaviors in HV-AV interactions using fixed reaction delays for human drivers. While these models offer a moderate representation of human driving patterns, their performance is limited by their limited parameter constraint, hindering their capacity to capture the intricate and complex behaviors exhibited by human drivers \cite{guo2020}. To overcome these limitations, various learning-based techniques have been explored \cite{di2021survey}. These include artificial neural networks such as multilayer networks, recurrent neural networks, and radial basis function networks \cite{morton2016}, as well as other methods like hidden Markov models \cite{qu2017} and Gaussian mixture regression \cite{lefevre2014a}. These approaches demonstrate increased precision in predicting human driver behavior. In a notable recent study \cite{huang2020learning}, the reaction time diversity of human drivers is addressed through the use of reinforcement learning to develop an adaptive optimal control for connected autonomous vehicles (CAVs) without estimating the driver behavior model. This methodology enhances the responsiveness and safety of the CAVs in mixed traffic by effectively adapting to varying platoon dynamics by explicitly considering the driver reaction times, yet it does not account for the uncertainty inherent in HVs in its adaptive optimal control design. Even though these methods offer enhanced accuracy in behavioral prediction, they do not adequately address modeling uncertainties as interpretable variances. This limitation poses challenges in incorporating uncertainties into control strategies and performing safety analyses. On the other hand, Gaussian process (GP) models, a nonparametric method extensively utilized in the modeling of complex robotic systems for learning-based control, offer benefits such as capturing complex behaviors with fewer parameters and providing interpretable uncertainties for performance analysis and safety assurance in control systems \cite{hewing2019}.

Recently, the application of GP models in physical human-robot interaction modeling has been explored. In \cite{haninger2022}, GP models were deployed to comprehend human force in varying task modes during a collaborative assembly involving a robotic manipulator, and an MPC approach was developed to decide the robot trajectory considering the inferred mode and task/human-associated objectives. Although this approach streamlined human-robot collaboration and boosted assembly efficiency by eliminating the need for mode-specific objective functions, it did not use the quantified modeling uncertainties of human force to augment the safety of human-robot interaction. It is critical to incorporate these uncertainties for effective control of human-robot interaction, however, given the limitations in current AV-HV modeling methods, we propose a hybrid model containing both a parametric model (nominal model) and a GP-based learning component \cite{wang2024improving}, encapsulating both deterministic and stochastic elements of human driver behavior. By including the GP component to learn discrepancies between the nominal model outputs and actual system behaviors \cite{wang2023learning}, and utilizing the uncertainties within the HV model in our MPC policy design, the safety of mixed-vehicle platooning control can be enhanced.

In the realm of MPC applications for mixed-traffic environments, a variety of innovative approaches have been introduced, each targeting specific challenges within this complex domain. The approach in \cite{guo2021anticipative} utilizes inverse MPC to enhance the anticipation of HV states, particularly in scenarios constrained by communication limitations. This method shows advancements in predicting vehicle behavior in mixed traffic. Meanwhile, the research in \cite{feng2021robust} introduces a Tube MPC framework that employs probabilistic bounds to manage uncertainties associated with HDV behavior, thereby ensuring robust control strategies in mixed traffic scenarios. Another noteworthy contribution is from \cite{zhan2022data}, which adopts a data-driven approach using Koopman operator theory. This method effectively represents mixed vehicle platoons by a linear model in a high-dimensional space, achieved through a neural network framework, and addresses the platoon control problem with both centralized and distributed MPC algorithms. Each of these methodologies offers unique solutions, addressing different facets of control and optimization in mixed traffic systems. 

Due to the major challenge in mixed traffic is how to handle the prediction uncertainty of HVs \cite{feng2021robust}, the major contribution of this paper is focused on proposing a novel GP-based HV model to quantitatively access the HV uncertainty and developing an MPC framework utilizing the HV model (GP-MPC), where the prediction uncertainty of HVs is handled explicitly. Compared with other state-of-the-art methods based on MPC, our framework has the following advantages. Firstly, the GP-MPC method employs GP learning, which enables direct and dynamic estimation of uncertainties in real-time traffic scenarios. This approach contrasts with methods such as those in \cite{guo2021anticipative}, which rely on inverse modeling techniques, and \cite{feng2021robust}, which needs to define probabilistic uncertainty bounds, by offering more adaptive and data-driven uncertainty management. Secondly, our GP-MPC framework enhances computational efficiency by leveraging the sparse GP model and the dynamic GP prediction within the MPC strategy, facilitating real-time applications in mixed-traffic environments. This is a significant improvement over the high-dimensional linear models used in \cite{zhan2022data}, which demand greater computational resources. Lastly, the GP-MPC provides a more flexible and comprehensive solution for handling the unpredictable and diverse behaviors of HVs, a crucial aspect that is not explicitly focused on in the other referenced methods. To the best of our knowledge, it is the first time that the quantified uncertainty of HV is explicitly used as a constraint in MPC policy design for platoon control in mixed traffic flow.

In our earlier work \cite{wang2024improving}, we validated the efficacy of our approach (GP-MPC) in enhancing HV modeling accuracy and boosting the safety of a mixed vehicle platoon. Nevertheless, the computational time required for each step of the GP-MPC was approximately 19 seconds, thereby curtailing its applicability for AV platoon control. To address this challenge, we have made significant advancements in our present work. We have successfully employed a sparse GP method \cite{snelson2005} and integrated dynamic sparse GP prediction into the MPC \cite{hewing2019}, as detailed in Sections \ref{section:sparse_gpr} and \ref{sec:dynamic_sparseGP_mpc}. These developments have led to a significant reduction in computational time by approximately 100 times, while simultaneously achieving considerable improvements in overall system control performance, as showcased in Sec. \ref{sec:simulations}. Additionally, another key limitation of our preliminary work \cite{wang2024learning-based} was the reliance on data obtained from a human-in-the-loop simulator, which falls short of representing real-world driving conditions comprehensively. To counter this limitation, we have collected data from field experiments in our present work, as elaborated in detail in Section \ref{sec:field}. This method enhances the quality of our model and extends its applicability to real-world scenarios.

The remaining sections of the paper are organized as follows. Sec. \ref{sec:HV_modeling} explains our novel HV modeling method. Sec. \ref{sec:field} describes the field experiments conducted for the HV modeling data collection. Sec. \ref{sec:controller} develops an MPC policy utilizing the proposed HV model. Sec. \ref{sec:simulations} presents extensive simulation tests to compare our developed MPC with a baseline MPC strategy. Finally, Sec. \ref{sec:conclusion} provides final thoughts and conclusions.

\section{Learning-based Human-driven Vehicle Model}
\label{sec:HV_modeling}
This section presents our approach to modeling human-driven vehicles (HVs). We first employ a time-delayed transfer function to capture human reaction time delay in an Auto-Regressive with Exogenous input (ARX) model. We then enhance this model with a GP component, improving prediction accuracy and quantifying uncertainty. To manage computational complexity, we applied the fully independent conditional (FIC) approximation to obtain a sparse GP model.

\subsection{First-principles Human-Driven Vehicle Model}
\label{sec:first_principles}
Reaction time delay is a notable human factor that influences operator performance \cite{pirani2022}. To accommodate the attributes of human central nervous system latencies, the neuromuscular system, and other human and environmental aspects, a time-delayed transfer function was developed in \cite{macadam2003} to characterize human responses:
\begin{equation} 
    G_{H}(s) \approx K \frac{1+T_{z} s}{1+2 \gamma T_{w} s+T_{w}^{2} s^{2}} e^{-T_{d} s}=\frac{\dot{P}_{H}(s)}{\dot{P}_{N}(s)}  \, . \tag{1} \label{eqn:TF_func}
\end{equation}
Here, $G_{H}(s)$ is the transfer function in the s-domain. The factor $K$ denotes the system's gain, indicating the proportional strength of the system's response. The term $T_{z}$ represents the zero time constant, $\gamma$ signifies the damping coefficient, and the variable $T_{w}$ indicates the system's natural frequency. The exponential term $e^{-T_{d} s}$ introduces a time delay in the system, with $T_{d}$ specifying the delay in the human driver's response. Lastly, $\dot{P}_{H}(s)$ and $\dot{P}_{N}(s)$ represent the Laplace transformations of the velocities for the HV $v_{k}^{H}$ and the front AV $v_{k}^{N_a}$, respectively. 

The parameters of $G_{H}(s)$ can be uncovered through the gathered data. Advancing our mathematical examination, we employ a second-order Padé approximation to the time delay element embedded within the transfer function \eqref{eqn:TF_func}. This usage enables us to infer a discrete-time difference equation for the ARX. The equation is derived by discretizing the transfer function using techniques outlined in \cite{wang2024improving}. The resultant equation is expressed as:
\begin{ceqn} 
    \begin{align}
        v^{H}_{k} &= -c_1 v^{H}_{k-1} - c_2 v^{H}_{k-2} - c_3 v^{H}_{k-3} - c_4 v^{H}_{k-4} \nonumber \\
        & \, \quad + b_1 v^{N_a}_{k-1} + b_2 v^{N_a}_{k-2} + b_3 v^{N_a}_{k-3} + b_4 v^{N_a}_{k-4} \, , \nonumber \\
        & = {f}\left(v_{k-1:k-4}^{H}, {v}_{k-1:k-4}^{N_a} \right) \, . \tag{2} \label{eqn:arx}
    \end{align}
\end{ceqn}
Here, $v^{H}_{k-i}$ and ${v}^{N_a}_{k-i}$ represent the velocities of the HV and the last vehicle in the AV platoon at time step $k-i$, respectively.

\subsection{The Proposed Human-Driven Vehicle Model}

Given the capability of Gaussian process (GP) models to enhance prediction accuracy and provide uncertainty estimations, they can be utilized as valuable constraints to provide safety guarantees in control systems. In our HV model, we leverage GP models to correct the predictions made by the ARX nominal model \eqref{eqn:arx}. However, it is important to note that the system state, represented by $v^{H}_{k}$ in \eqref{eqn:arx}, does not adhere to the common assumption of the Markov property in learning-based control system models. To address this challenge, we propose a novel approach for modeling HVs using an ARX+GP format, which entails employing two separate equations as:
\begin{subequations}
\label{eq:arx_gp_model}
\begin{align}
    v^H_k &= \sum_{i=1}^4 -c_i v^H_{k-i} + \sum_{i=1}^4 b_i v^{N_a}_{k-i}  = {f}(\cdot) \quad \text{ (see \ref{eqn:arx})} \, , \tag{3a} \label{eqn:HV_nominal}\\
    \tilde{v}^{H}_{k}&=\overbrace{v^H_k}^{\text {ARX prediction}}+\overbrace{{g}(v^{H}_{k-1}, v^{N_a}_{k-1})} ^{\text {GP-based correction}} \, . \tag{3b} \label{eqn:HV_corrected_modeling}
\end{align}
\end{subequations}
In our proposed framework, the GP-compensated velocity prediction of the HV is denoted as $\tilde{v}^{H}_{k}$. The function ${f}(\cdot)$ serves as an   ARX nominal model with constant parameters $b_i$ and $c_i$. Meanwhile, the GP model ${g}(\cdot)$ learns the discrepancies between the ARX model and the actual system behaviors, derived from the observed system data. It's worth noting that ${g}(\cdot)$ relies on both the HV velocity $v^{H}$ and the last AV's velocity $v^{N_a}$, which ensures it effectively captures the HV dynamics influenced by the leading AV's actions.

\subsection{GP Model Training}
\label{section:sparse_gpr}

In equation \eqref{eqn:HV_corrected_modeling}, the GP model is used to compensate for discrepancies between the actual and the ARX-predicted velocities of HVs. By utilizing prior velocity state data, $v^{H}_{j}$ and $v^{N_a}_{j}$, the discrepancies can be learned by GP models as:
\begin{equation}
    \hat{v}^{H}_{j} -{v}^{H}_{j} = {g}({a}_{j-1}) \, .  \tag{4} \label{eqn:gp_data_prep}
\end{equation}
In this equation, ${a}_{j-1}=(v^{H}_{j-1}, v^{N_a}_{j-1})$ denotes the discrepancy state. $\hat{v}^{H}_{j}$ and $v^H_j$ refer to velocities obtained from collected data and predicted by the ARX model in \eqref{eqn:HV_nominal}, respectively. Here, ${a}_{j}$ symbolizes an individual data point within the observed input-output dataset.

In each experiment, all input-output data (one-dimensional) pairs are gathered at every time step. These data are then formatted according to equation \eqref{eqn:gp_data_prep} to create a discrepancy dataset, denoted as $\mathcal{D} = \{\mathbf{a} = [{a}_1, \cdots, {a}_n]^\top \in \mathbb{R}^{n_a \times n}, \mathbf{g} = [{g}_1, \cdots, {g}_n]^\top \in \mathbb{R}^{1 \times n}\}$, where $n$ stands for the total number of data in the experiment. In line with many robotics control methodologies using GP techniques \cite{hewing2019}, a squared exponential kernel was chosen as the kernel/covariance function for this work:
\begin{ceqn}
    \begin{equation} 
        k({a}_i, {a}_j) = \sigma_{f}^2 \exp \left[-\frac{1}{2}
        \left({a}_i - {a}_j \right)^\top L^{-1}
        ({a}_i - {a}_j) \right] \, , \tag{5} \label{eqn:rbf}
    \end{equation}
\end{ceqn}
where ${a}_i$, ${a}_j \in \mathbf{a}$. The hyperparameters $\sigma_{f}^2$ and $L \in \mathbb{R}^{n_a \times n_a}$ are optimized during the training process of the GP models by maximizing the log marginal likelihood of the observed dataset \cite{wang2023intuitive}.

The GP prior model has a zero mean ${\mu}(\cdot)$ and a prior kernel $k(\cdot, \cdot)$. Conditioning on the observed data $\mathcal{D}$ results in a predictive posterior distribution, the unknown function ${g}(\cdot)$ is approximated as a trained GP model $d(\cdot)$ that can be used to make predictions at a new data point ${a}^* \in \mathbf{a}^*$ \cite{rasmussen2006} through
\begin{ceqn} 
    \begin{equation} 
        {g}({a}^*) \approx {d}({a}^*) \sim \mathcal{N} \left({\mu}^{{d}}({a}^*), {\Sigma}^{{d}}({a}^*) \right) \, ,  \tag{6} \label{eqn:gp_eqn}
    \end{equation}
\end{ceqn}
where 
\begin{ceqn}
    \begin{align} 
       \mu^d({a}^*) &= \mathbf{K}_{{{a}^*}\mathbf{a}} \left(\mathbf{K}_{\mathbf{a}\mathbf{a}} + \sigma^2 \mathbf{I} \right)^{-1} \mathbf{g} \, , \tag{7a} \label{eqn:gp_mean}   \\ 
       \Sigma^d({a}^*) &= \mathbf{K}_{{{a}^*}{{a}^*}} -  \mathbf{K}_{{{a}^*}\mathbf{a}}\left(\mathbf{K}_{\mathbf{a}\mathbf{a}} + \sigma^2 \mathbf{I} \right)^{-1} \mathbf{K}_{\mathbf{a}{{a}^*}} \, . \tag{7b} \label{eqn:gp_var}
    \end{align}
\end{ceqn}
The observed data $\mathbf{g}$ follows a normal distribution $\mathbf{g} \sim \mathcal{N} \left(0, \mathbf{K}_{\mathbf{a}\mathbf{a}} + \sigma^2 \right)$. Here, $\mathbf{K}_{\mathbf{a}\mathbf{a}}$ denotes the Gram matrix obtained by applying the kernel function $k(\cdot, \cdot)$ to the observed input data $\mathbf{a}$, i.e., $[\mathbf{K}_{\mathbf{a}\mathbf{a}}]_{ij} = k({a}_i, {a}_j)$. Moreover, $[\mathbf{K}_{\mathbf{a}{a}^*}] = k({a}_j, {a}^*)$, $\mathbf{K}_{{{a}^*}\mathbf{a}} = (\mathbf{K}_{\mathbf{a}{{a}^*}})^\top$, and $\mathbf{K}_{{{a}^*}{{a}^*}} = k({a}^*,{a}^*)$. 

The computational complexity of the mean \eqref{eqn:gp_mean} and variance \eqref{eqn:gp_var} have complexities of $\mathcal{O}(n_a m)$ and $\mathcal{O}(n_a m^2)$ respectively, scaling with the total number of training data points $m$ \cite{hewing2019}. To maintain a balance between computational demand and approximation accuracy, we applied the fully independent conditional (FIC) approximation \cite{snelson2005}, a state-of-the-art sparse spectrum method, to the standard GP model \eqref{eqn:gp_eqn}. This method condenses the information from the training data into a subset called inducing variables, only requiring the storage of this subset rather than the entire dataset. In the FIC, the locations of these inducing variables are automatically determined by maximizing the GP marginal likelihood by gradient ascent.

\subsection{Assessing Traditional Car-Following Models}
Conventional car-following models like the IDM, OVM, and Gipps' model surveyed in Sec. \ref{sec:related_work}, which are developed from extensive human-driving datasets, primarily focuses on typical behavior patterns of human drivers. These models, while effective in representing general driving behavior, are primarily deterministic or static in their probabilistic approach as the nominal model \eqref{eqn:arx}. 

Our GP-based HV model, however, diverges significantly from these traditional models by leveraging Gaussian processes (GP) to capture and quantify the uncertainties and variations inherent in human driving behavior. This approach provides a flexible and probabilistic understanding of driving behaviors, crucial for scenarios with significant deviations from average patterns. In contrast to traditional models that rely on predefined rules or parameters, our model incorporates a GP learning-based component. This allows for adaptive learning from data, offering a nuanced representation of human-driven vehicle behavior, which is particularly beneficial in mixed-traffic environments where human behavior can be highly unpredictable. While traditional models and our approach share the commonality of focusing on driver behavior and vehicle interactions, our model enhances this paradigm by integrating machine learning techniques. This results in a more accurate reflection of the complexities of human driving behavior, a critical advancement over the more deterministic formulations of conventional car-following models. 

\section{Field Experiment}
\label{sec:field}
This section details the settings of the field experiments we conducted for data collection. These include descriptions of the test track, vehicles, drivers, and an introduction to the object tracker on the autonomous driving system that contributes directly to the velocity estimation, which is critical data for training our proposed model. 

\subsection{Test Track}
\begin{figure}
    \centerline{\includegraphics[width=0.92\columnwidth]{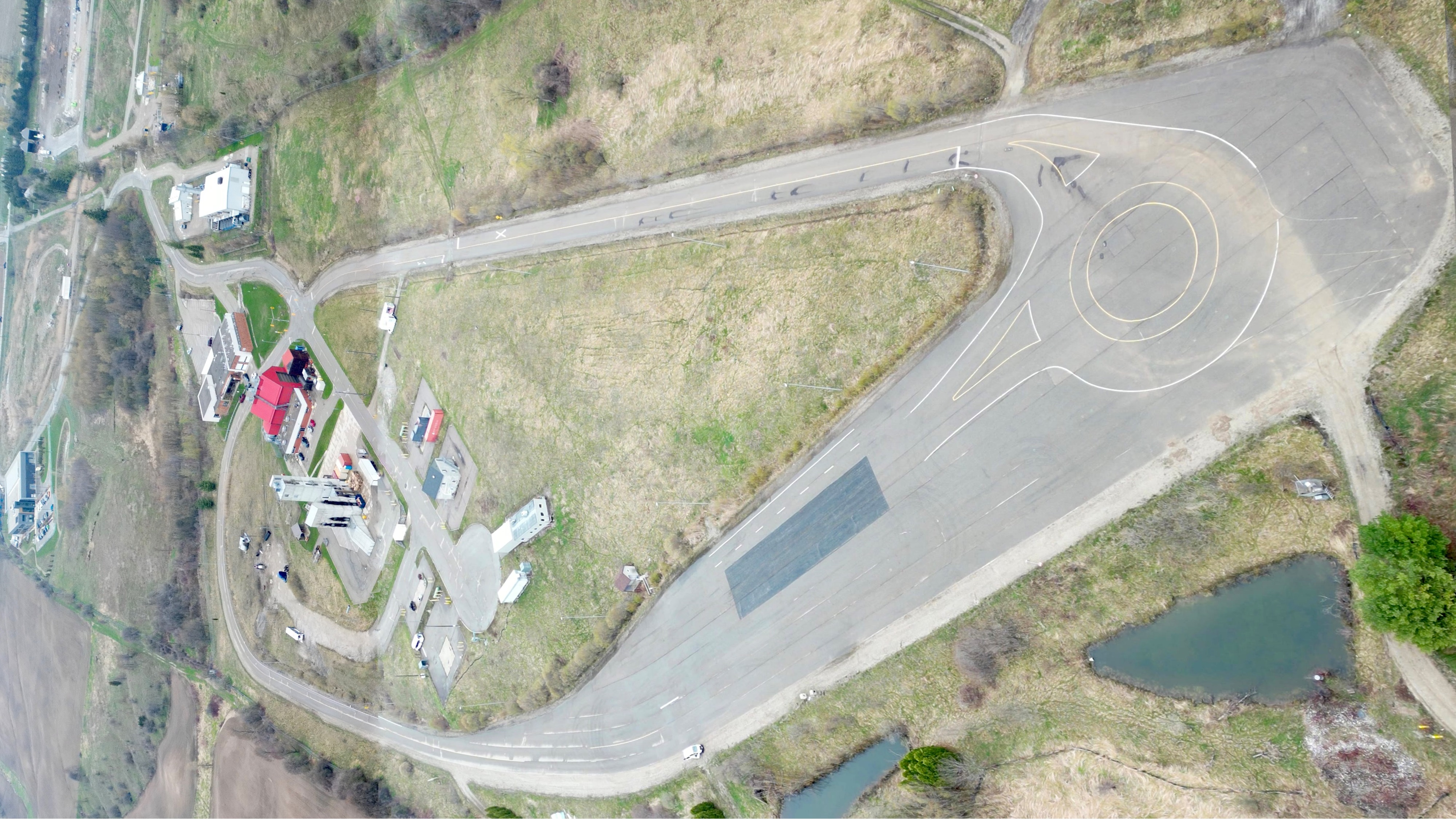}}
    \caption{An aerial image of the test track to conduct the field experiments of a human-driven vehicle following an autonomous vehicle. 
    }
    \label{figure:dji_all_track_paths}
\end{figure}

\begin{figure}
    \centerline{\includegraphics[width=0.92\columnwidth]{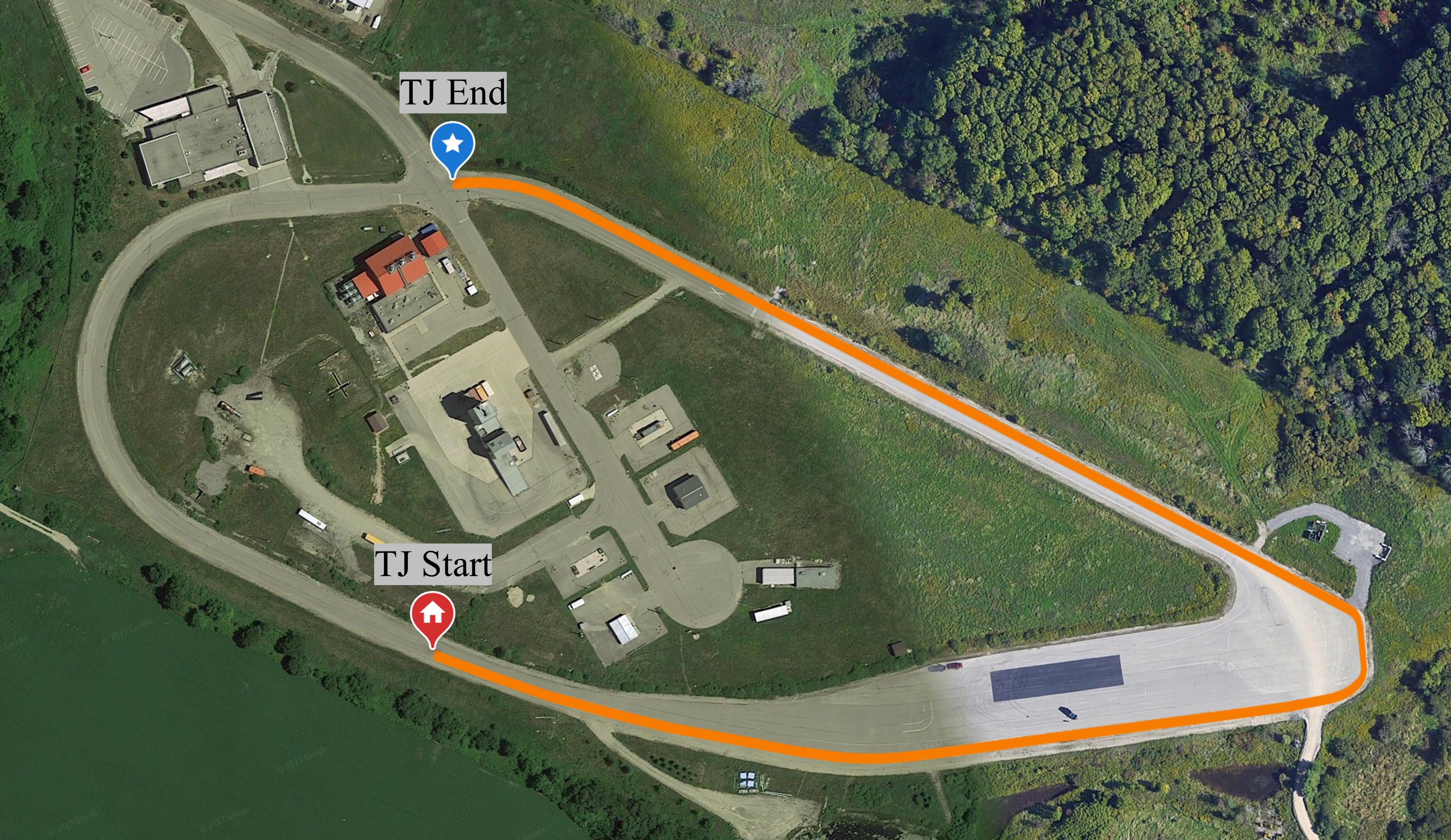}}
    \caption{The path for the field experiments shown in a GPS photo of the field test track.
    }
    \label{figure:gps_track_paths}
\end{figure}

We performed the field experiments involving an HV following an AV at the Waterloo Region Emergency Services Training and Research Center in Waterloo, Ontario, Canada, shown in Fig. \ref{figure:dji_all_track_paths}. We gathered data on an approximately 850-meter stretch of the track, featuring varying curvatures and bumps to simulate realistic traffic jam (TJ) scenarios. This stretch is highlighted by the orange line in Fig. \ref{figure:gps_track_paths}. The car-following experiment starts from the ``TJ Start'' point and ends at the ``TJ End''.


\subsection{Test Vehicles and Drivers}
\begin{figure}
    \centerline{\includegraphics[width=0.92\columnwidth]{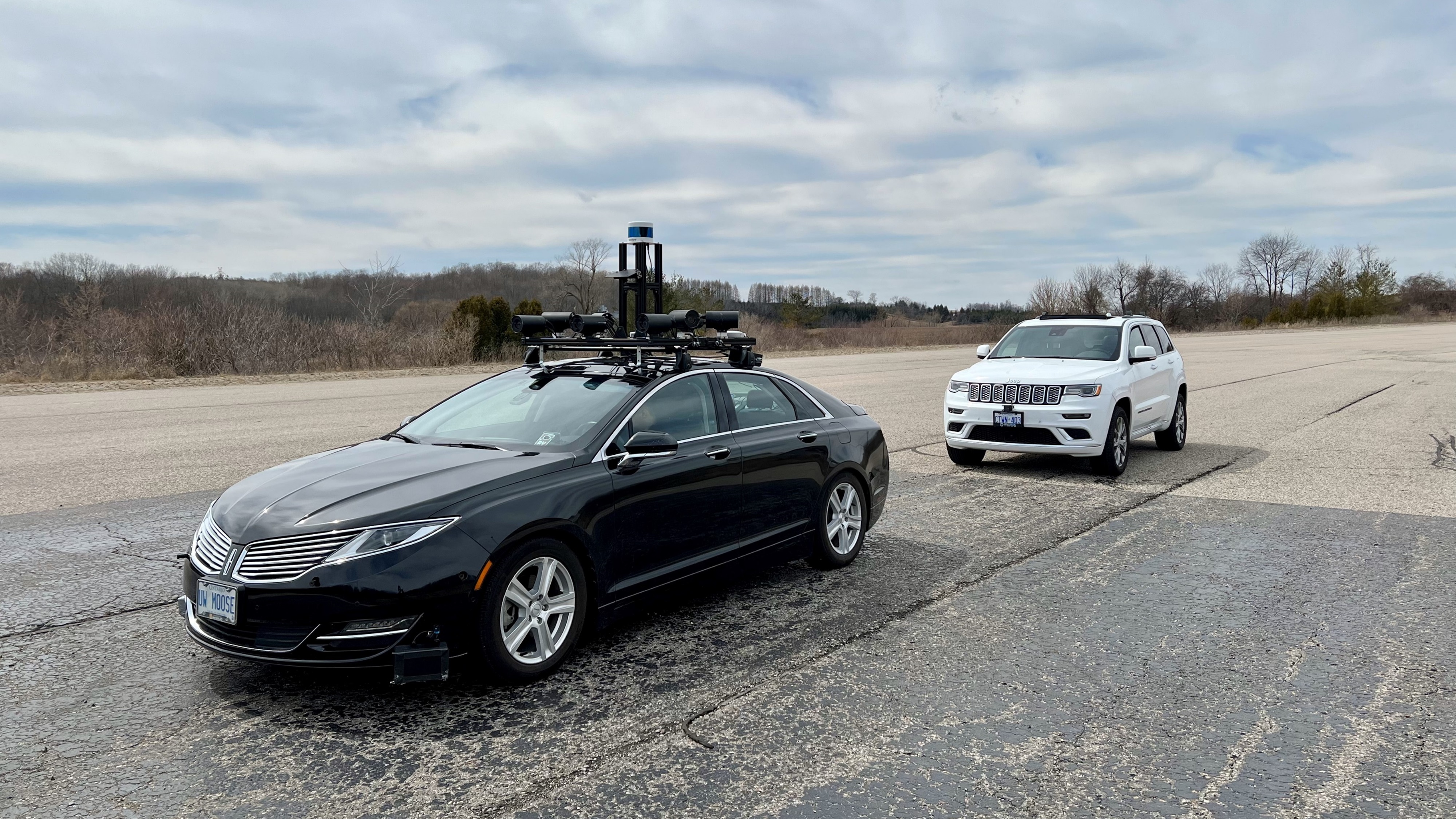}}
    \caption{The autonomous vehicle platform, UW Moose, and a following human-driven vehicle on the test track.
    }
    \label{figure:AV_HV_cars}
\end{figure}

We utilized the ``UW Moose'' platform \cite{wiseadsmooseweb,pitropov2021} shown in Fig. \ref{figure:AV_HV_cars} as our AV, which was developed at the University of Waterloo and which was the first Canadian-built automated driving system tested on public roads in Canada in August 2018. The platform is a modified 2015 Lincoln MKZ Hybrid that has been converted to an autonomous drive-by-wire operation. The UW Moose platform is equipped with a full suite of sensors, including a LiDAR, cameras, GNSS, and IMU, which are used to gather comprehensive data about the vehicle platform and its surrounding environment. The platform runs a variety of software modules for perception, planning, and control, which are responsible for processing sensor data, generating maps of the environment, planning safe and efficient trajectories, and controlling the vehicle's motion~\cite{wiseadsmooseweb}. The platform has been tested in various driving scenarios in urban environments, and it has demonstrated the ability to operate at limited automation Level 3. 

To devise the HV model using our proposed method, we required velocity data from both the HV and AV as per equations \eqref{eqn:HV_nominal} and \eqref{eqn:HV_corrected_modeling}. We focus on introducing the software components and sensors that are directly related to this velocity estimation purpose, leveraging the UW Moose platform's sophisticated sensor suite and software algorithms for autonomous navigation and control. The software stack includes a localizer, a 3D object detector, and a Kalman filter (KF) tracker. These components collaborate to estimate the platform's velocity and the velocity of surrounding objects. The localizer uses GNSS, IMU, and wheel encoders data to optimally estimate the platform's position and orientation at 100\,Hz, while the 3D object detector leverages Velodyne VLS-128 AlphaPrime lidar and OpenPCDet package \cite{openpcdet2020} to conduct 3D object detection at 20\,Hz. The KF tracker improves velocity estimate accuracy by utilizing data from the localizer and 3D object detector at 20\,Hz.

\begin{figure}
    \centering
    \vspace{0.15cm}
    \includegraphics[trim=0cm 0cm 0cm 0cm, width=0.99\columnwidth]{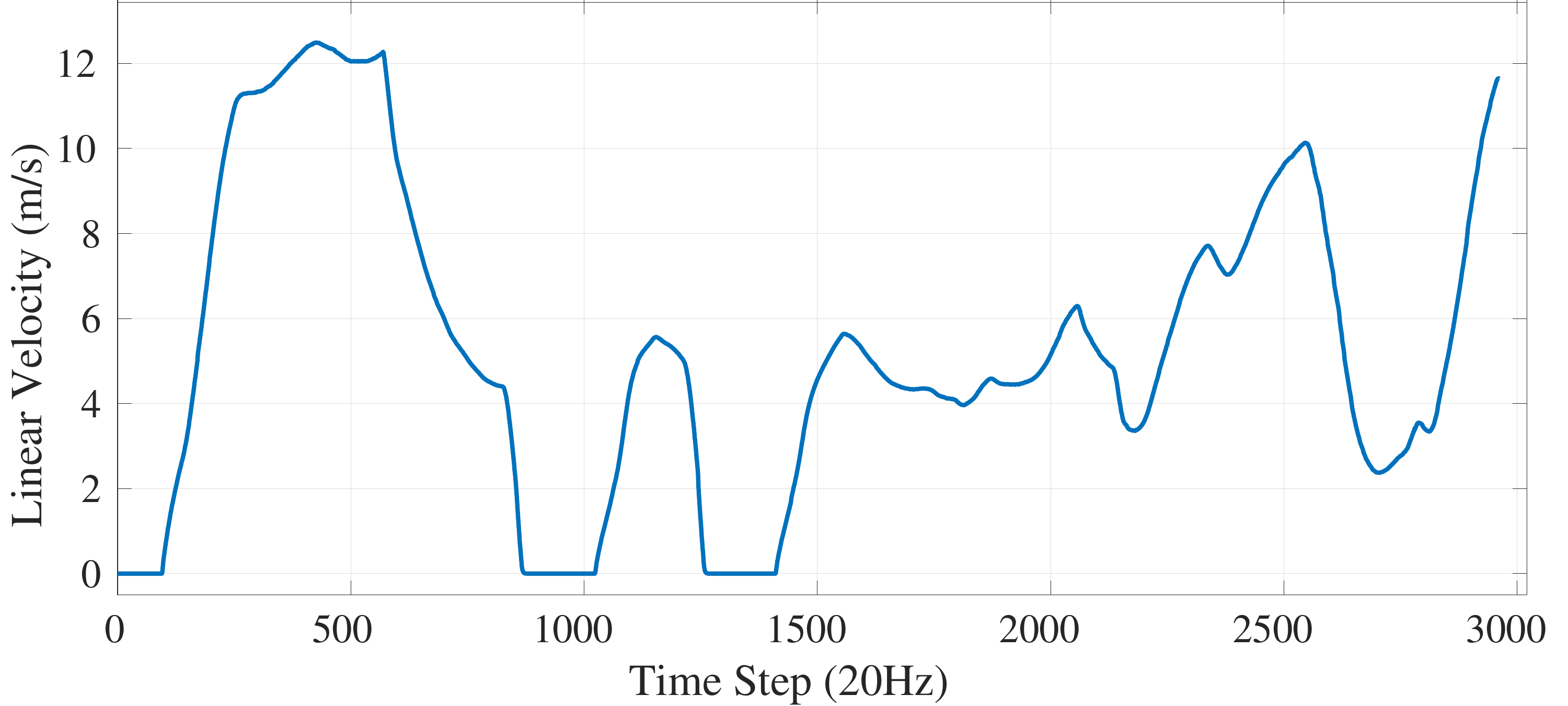}
    \caption{The pre-defined velocity profile for the autonomous vehicle to follow.
    }
    \label{figure:AV_velocity_profile}
\end{figure}

For the HV, we chose a Jeep Grand Cherokee operated by six drivers from our research lab, with diverse driving behaviors to ensure a broad range of data. We selected drivers based on gender, driving experience, and familiarity with AV dynamics. Specifically, there are 1 female and 5 male drivers, the driving experience ranging from 1 to 11 years, and 3 out of 6 are familiar with AV dynamic behaviors. Each driver followed the AV twice in each experiment. The AV localized itself using a pre-defined HD map of the test track and followed a simulated vehicle following a pre-defined velocity profile to simulate real traffic jam scenarios. The test was conducted in mixed-reality mode, where the follower AV (UW Moose) followed a simulated AV \cite{antkiewicz2020}. The velocity profile for the simulated AV is shown in Fig. \ref{figure:AV_velocity_profile}. 

\section{Modeling of Human-Driven Vehicle }
\label{sec:HV_model}

In the proposed GP+ARX model for HVs, represented in equation \eqref{eqn:HV_corrected_modeling}, we estimate the ARX model using data collected from a human-in-the-loop simulator (see Fig. \ref{figure:car_simulator}). For the training of the GP model, we adhere to the methodology specified in Sec. \ref{section:sparse_gpr}, applying data collected during the field experiment described in Sec. \ref{sec:field}.

\subsection{ARX Nominal Model}
\begin{figure}
    \centering
    \vspace{0.15cm}
    \includegraphics[trim=0cm 0cm 0cm 0cm, width=0.92\columnwidth]{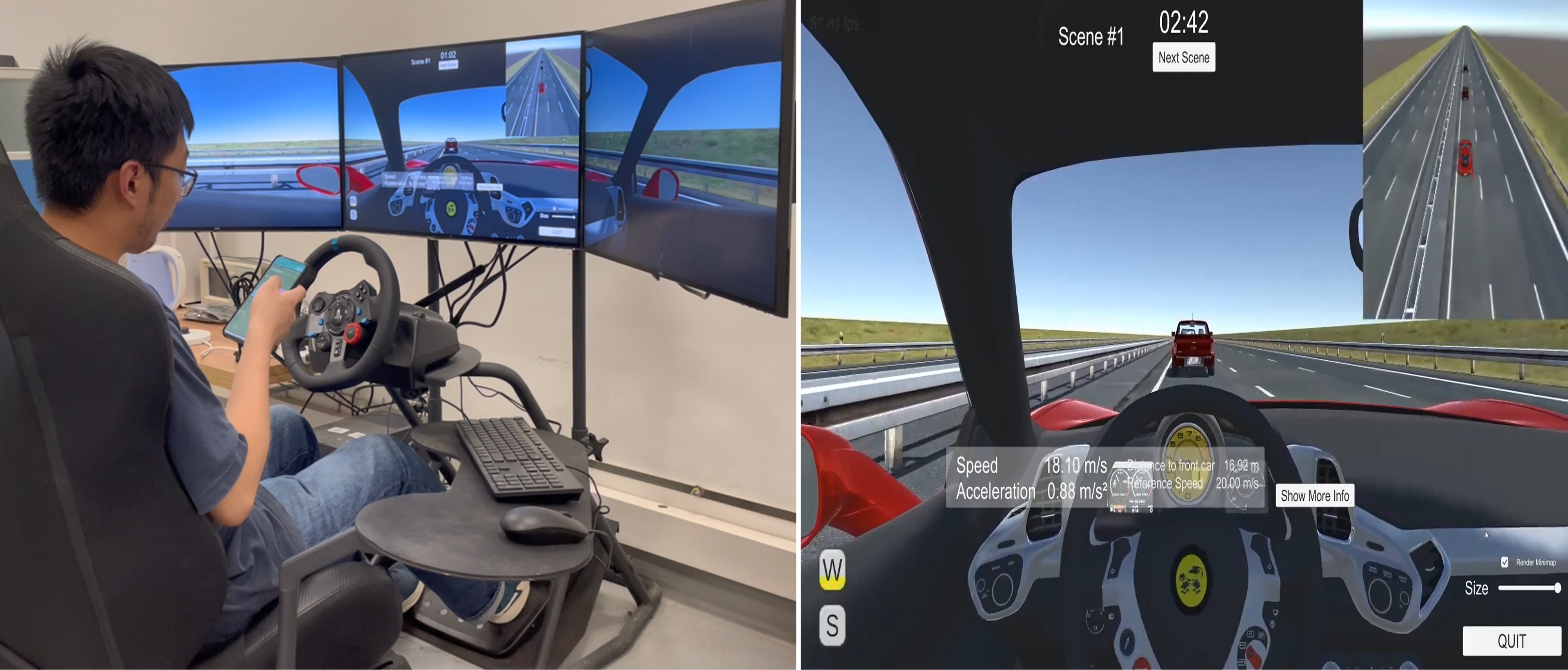}
    \caption{Data was gathered through a Unity driving simulator, featuring three individual drivers under the influence of distractions across three varied driving scenarios. In every experiment, the human driver was tasked to follow a platoon of two AVs.
    }
    \label{figure:car_simulator}
\end{figure}

The identification of the ARX nominal model for HVs makes use of the data collected from three unique driving scenarios. Three drivers followed a platoon of two AVs within a Unity-based simulated environment, as shown in Fig. \ref{figure:car_simulator}.
During the experiments, we subjected the drivers to cognitive distractions by providing them algebraic problems with multiple choices to solve while they followed the AV platoon \footnote{See \url{https://youtu.be/mfPzYDQrvV4} for video playback.}. The modeling of such distracted driving scenarios is crucial due to the added uncertainty they introduce to the HV model, making maintaining safe control in AV-HV interactions more challenging. Each experiment required the driver to follow the AV platoon for three minutes. The lead AV adhered to a reference velocity profile divided into three segments: a constant 20\,m$/$s for 30 seconds, a linear chirp waveform from 25 to 10\,m$/$s over 60 seconds, and a constant 10\,m$/$s for 90 seconds \cite[Sec. V.B]{pirani2022}. We calculated mean values from the collected data points, which subsequently were utilized for the derivation of the transfer function per equation \eqref{eqn:TF_func}. By referencing equation \eqref{eqn:arx}, we identified the ARX nominal model parameters in \eqref{eqn:arx} as $c_1 = -3.0227$, $c_2 = 3.3543$, $c_3 = -1.6329$, $c_4 = 0.3014$, $b_1 = 0.0063$, $b_2 = -0.0303$, $b_3 = 0.0495$, and $b_4 = -0.0254$.

\subsection{Sparse GP+ARX Model}
\label{sec:spares_gp_arx}

To estimate the discrepancy between the ARX nominal model's prediction and the actual HV behavior, we prepared the field experiment data using equation \eqref{eqn:gp_data_prep}. During the field experiments, we collected 12 data sets in total, consisting of 2 datasets for each driver. We carefully selected one set from each driver as training data (6 sets in total) and remained another 2 sets for testing. The primary selection criterion was the time delay of the object tracker on the UW Moose AV platform. We preferred datasets with a consistent time delay of less than 0.1 seconds for the HV velocity estimation, as they yielded more accurate results.

To reduce the computational burden of employing the ARX+GP model in the subsequent mixed vehicle platooning control policy design, we trained the GP model using around 20\% of the data points randomly selected from the six training datasets. We used this approach because the prediction time of the GP model increases significantly with the number of training data points, as discussed in Sec. \ref{section:sparse_gpr}. To achieve a balance between computational efficiency and model accuracy, we applied the FIC sparse approximation technique to the trained GP model, thereby speeding up the prediction process. The hyperparameters for the sparse GP model were inherited from the trained standard GP model, and 20 inducing points were automatically selected by the FIC method within the training datasets. Remarkably, the FIC sparse GP model demonstrated an impressive reduction in prediction time, taking only 0.00021 seconds on average compared to the standard GP model's 0.0037 seconds per prediction. This signifies an impressive 18-fold increase in prediction speed for the FIC sparse GP model over the standard GP model.

\begin{figure}
    \centering
    \vspace{0.15cm}
    \subfloat{{\includegraphics[trim=0cm 0cm 0cm 0cm, width=0.98\columnwidth]{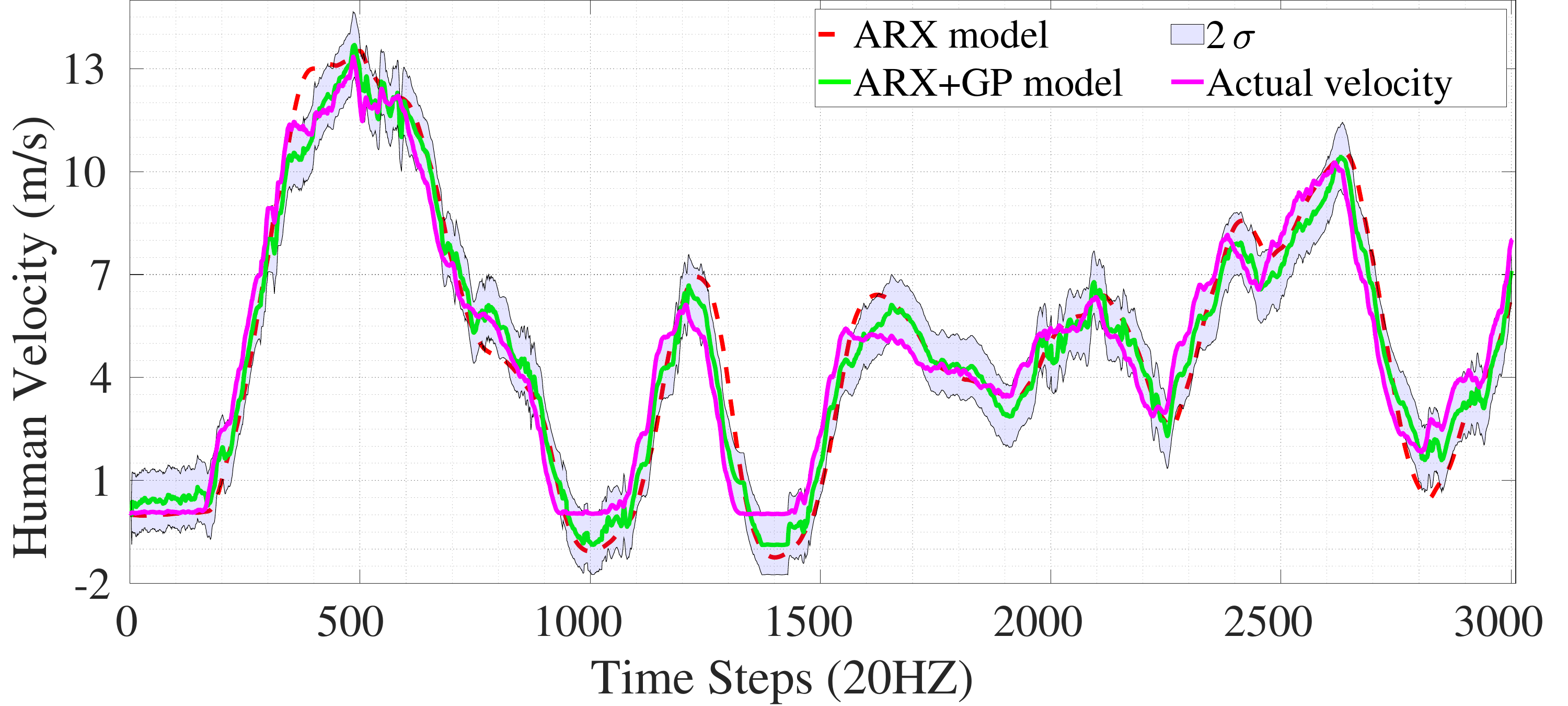} }}
    \qquad 
    \vspace{0.02cm}
    \subfloat{{\includegraphics[trim=0cm 0cm 0.0cm 0.0cm, width=0.98\columnwidth]{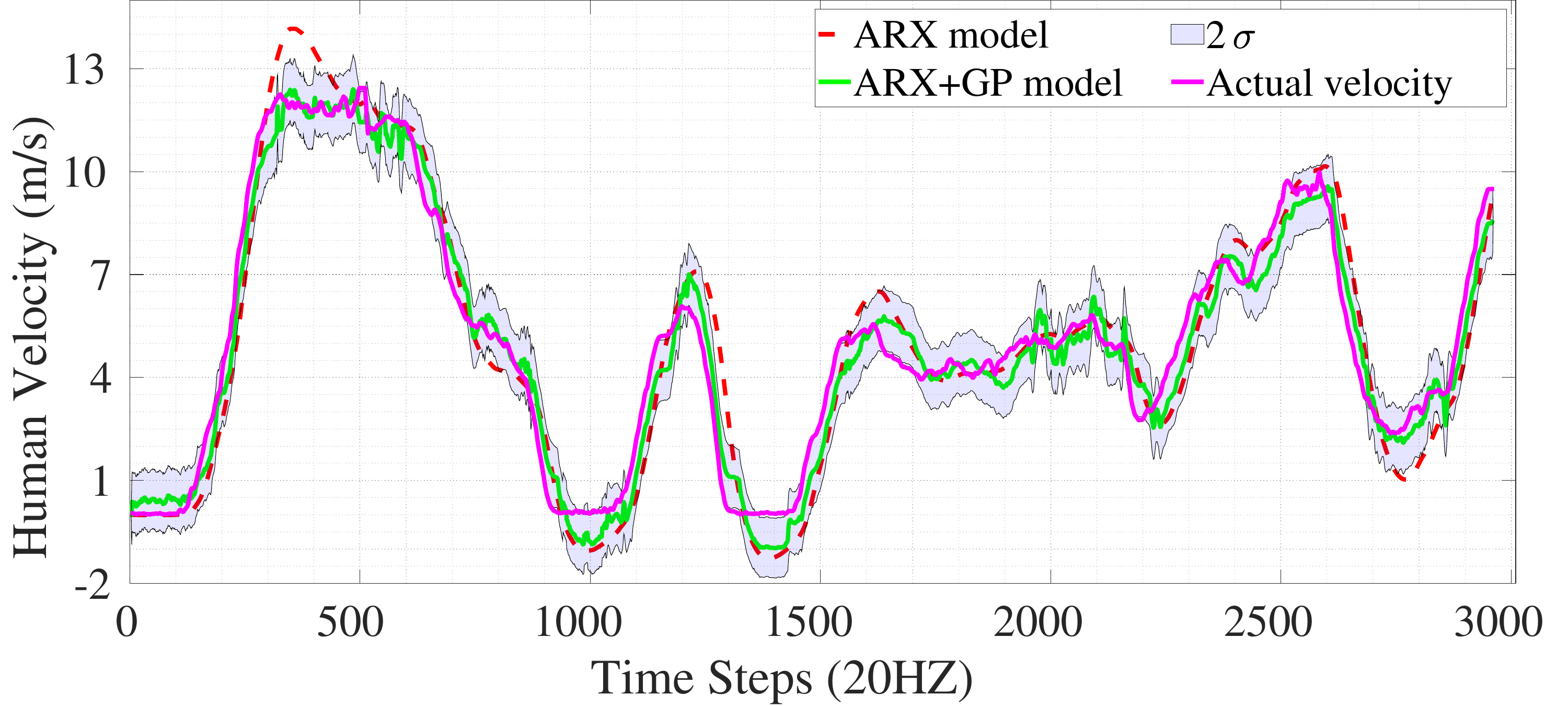} }}
    \caption{The GP model's performance was evaluated on two testing datasets. The velocity predictions by simulating the ARX model of \eqref{eqn:HV_nominal} and ARX+GP model of \eqref{eqn:HV_corrected_modeling} were compared, plotting them alongside the measured velocities and twice the standard deviation (2$\sigma$) from the GP model. The results highlight that the ARX+GP model considerably improved the fit of actual velocity curves compared to the standalone ARX model. On average, the ARX+GP model enhances modeling accuracy by 36.34\% in terms of RMSE compared to the ARX model.}
    \label{figure:gp_standard}
\end{figure}

To demonstrate the accuracy enhancement achieved by integrating the sparse GP model, we plotted the testing results for both sets of testing data. Fig. \ref{figure:gp_standard} demonstrates the velocity predictions of the ARX and ARX+GP models along with twice the standard deviation (2$\sigma$) estimated by the sparse GP model. We also included the actual measured HV velocities in the plots to emphasize  the accuracy improvement achieved with the GP model. We observed that the ARX+GP model aligned with the actual velocity data curves better than the ARX model. 
To quantify the accuracy improvement, we computed the root mean square error (RMSE) for each testing dataset using
\begin{ceqn}
    \begin{align}
        \text{RMSE}_{{v}^{H}_*}=\sqrt{\sum_{i=1}^n \frac{\left({v}^{H}_*-{v}^{H}_\text{act}\right)^2}{N}} \, , \tag{8} \label{eqn:rmse}
    \end{align}
\end{ceqn}
where ${v}^{H}_*$ represents either the ARX or the ARX+GP model's velocity predictions over $N$ samples in a testing dataset, and ${v}^{H}_\text{act}$ denotes the actual measured velocities. The RMSE quantifies the average magnitude of the prediction errors in the model across the testing datasets, hence lower RMSE values indicate a higher accuracy in modeling the HVs. The RMSE results for the ARX and ARX+GP models are summarized in Tab. \ref{tab:rmse_gp}. The ARX model yielded an average RMSE of 1.076, while the ARX+GP model achieved a significantly lower average RMSE of 0.685, representing an overall modeling accuracy improvement of approximately 36.34\% for the ARX+GP model compared to the ARX model. This enhanced modeling accuracy is visibly illustrated in Fig. \ref{figure:gp_standard}, where the ARX+GP model shows a closer alignment with the actual velocity data curves than the ARX model.

\begin{table}
    \caption{RMSE results of the ARX and sparse ARX+GP models with two testing datasets.}
    \label{tab:rmse_gp}
    \centering
    \begin{tabular}{|c|c|c|c|}
        \hline
        \textbf{Testing Dataset} & \textbf{Dataset 1} & \textbf{Dataset 2} \\
        \hline
        ARX model & 1.0522 & 1.1006 \\
        \hline
        ARX+GP model & 0.6794 & 0.6907 \\
        \hline
    \end{tabular}
\end{table}

\section{Controller Design}
\label{sec:controller}

Our proposed modeling approach's effectiveness for HVs is further demonstrated through a MPC strategy designed specifically to utilize this model. This demonstration highlights the effectiveness of the proposed approach in improving both performance and safety in the control of a mixed-vehicle platoon. The suggested method is particularly tailored for longitudinal car-following situations where an HV trails an AV platoon and does not apply to other AV-HV interactions, such as lane-changing maneuvers.

\subsection{System Model}

Visualize a mixed vehicle platoon consisting of AVs and a single HV, referred to as $A^{\mathbf{n_a}}$ and $H$, respectively. Here, $\mathbf{n_a} = \{1, 2, \cdots, N_a\}$, with $N_a$ signifying the quantity of AVs. At the present time step $k$, the velocity and position of $A^{\mathbf{n_a}}$ are symbolized by $p^{\mathbf{n_a}}_k$ and $v^{\mathbf{n_a}}_k$, respectively. Considering the kinematic model of AVs as follows:
\begin{ceqn}
    \begin{align}  
    v_{k+1}^{\mathbf{n_a}} &= v_{k}^{\mathbf{n_a}} + T \, \mathrm{acc}_{k}^{\mathbf{n_a}} \, , \tag{9a} \label{eqn:av_eqn_a} \\
    p_{k+1}^{\mathbf{n_a}} &= p_{k}^{\mathbf{n_a}} + T \, v_{k}^{\mathbf{n_a}} \, . \tag{9b} \label{eqn:av_eqn_b}
    \end{align}
\end{ceqn}
In the above, $0<T \ll 1$ indicates the sample time, and $\mathrm{acc}_{k}^{\mathbf{n_a}}$ signifies the acceleration of $A^{\mathbf{n_a}}$. We operate under the assumption that AVs are deterministic, measuring and communicating their states flawlessly, i.e., $\Sigma(v_{k}^{\mathbf{n_a}}) = 0$. Upon application of \eqref{eqn:HV_corrected_modeling}, the human-operated vehicle model becomes:
\begin{ceqn} 
    \begin{align} 
        \tilde{v}_{k}^{H} &= v_{k}^{H} + {g}(v_{k-1}^{H}, {v}_{k-1}^{N_a}) \, , \tag{10a} \label{eqn:sys_model_a} \\
        p_{k+1}^{H} &= p_{k}^{H} + T \, \tilde{v}_{k}^{H} \, . \tag{10b} \label{eqn:sys_model_b}
    \end{align}
\end{ceqn}
Here, $\tilde{v}_{k}^{H}$ represents the compensated HV velocity obtained from the GP model, while $v_{k}^{H}$ (computed using \eqref{eqn:arx}) and $p_{k}^{H}$ denote the velocity and position states at the current time step $k$ respectively, whose variances $\Sigma(v_{k}^{H})$ and $\Sigma(p_{k}^{H})$ are both equal to zero. 

Utilizing \eqref{eqn:sys_model_b} and \eqref{eqn:gp_eqn}, the propagation of the HV position mean $\mu \left(p_{k+1}^{H} \right)$ is derived as follows:
\begin{ceqn} 
    \begin{align} 
        \mu \left(p_{k+1}^{H} \right) &= \mu \left( p_{k}^{H} + T \, v_{k}^{H} + T \, {d}(v_{k-1}^{H}, {v}_{k-1}^{N_a}) \right) \, , \tag{11a} \label{eqn:mean_prop_a} \\
        &= \mu \left( p_{k}^{H} \right) + T \, \mu \left( v_{k}^{H} \right) + T \, \mu \left( {d}(v_{k-1}^{H}, {v}_{k-1}^{N_a}) \right) \, , \tag{11b} \label{eqn:mean_prop_b}  \\
        &= \mu \left( p_{k}^{H} \right) + T \, \mu \left( v_{k}^{H} \right) + T \, \mu^d \left( v_{k-1}^{H}, {v}_{k-1}^{N_a} \right) \, . \tag{11c} \label{eqn:mean_prop_c} 
    \end{align}
\end{ceqn}
To express the equation more concisely, we denote the HV position mean at time step $k+1$ on the left side of \eqref{eqn:mean_prop_c} as $\mu_{k+1}^{p^{H}}$, and correspondingly, we can rewrite the terms on the right side of the equation. Thus, equation \eqref{eqn:mean_prop_c} is compactly represented as:
\begin{ceqn} 
    \begin{align} 
        \mu_{k+1}^{p^{H}} &= \mu_{k}^{p^{H}} + T \, v_{k}^{H} + T \, \mu^d \left( v_{k-1}^{H}, {v}_{k-1}^{N_a} \right) \, , \tag{11d} \label{eqn:mean_prop_d} 
    \end{align}
\end{ceqn}
with the initial value $\mu_{0}^{p^{H}} = p_{k}^{H}$. By using \eqref{eqn:sys_model_b} and \eqref{eqn:gp_eqn}, we derive the propagation of the HV position variance, $\Sigma \left(p_{k+1}^{H} \right)$, as:
\begin{ceqn} 
    \begin{align} 
        \Sigma \left(p_{k+1}^{H} \right) &= \Sigma \left( p_{k}^{H} + T \, v_{k}^{H} + T \, {d}(v_{k-1}^{H}, {v}_{k-1}^{N_a}) \right) \, , \tag{12a} \label{eqn:variance_prop_a} \\
        & = \Sigma(p_{k}^{H}) + \Sigma(T \, v_{k}^{H}) + \Sigma \left( T \, {d}(v_{k-1}^{H}, {v}_{k-1}^{N_a}) \right) \, , \tag{12b} \label{eqn:variance_prop_b} \\
        & = \Sigma (p_{k}^{H}) + T^2 \Sigma^d \left(v_{k-1}^{H}, {v}_{k-1}^{N_a}\right) \, . \tag{12c} \label{eqn:variance_prop_c}
    \end{align}
\end{ceqn}
Here, at the current time step $k$, the variances $\Sigma(v_{k}^{H})=0$. Similarly, to express the equation more succinctly, we denote the HV position variance at time step $k+1$ on the left side of \eqref{eqn:variance_prop_c} as $\Sigma_{k+1}^{p^{H}}$, and correspondingly rewrite the terms on the right side of the equation. This allows us to compactly represent equation \eqref{eqn:variance_prop_c} as:
\begin{ceqn} 
    \begin{align} 
        \Sigma_{k+1}^{p^{H}} = \Sigma_{k}^{p^{H}} + T^2 \Sigma^d \left(v_{k-1}^{H}, {v}_{k-1}^{N_a}\right) \ . \tag{12d} \label{eqn:variance_prop_d}
    \end{align}
\end{ceqn}
Here, the initial value is $\Sigma_{0}^{p^{H}} = 0$, and it is important to note that the HV variance propagation does not consider the covariance between $p_{k}^{H}$ and $v_{k}^{H}$.

\subsection{Safe Distance Probability Constraint}
To maintain a secure distance within the mixed platoon of vehicles, a distance constraint is established between AVs, defined by a constant $\Delta$, such that $p_{k}^{\mathbf{n_a}-1}-p_{k}^{\mathbf{n_a}} > \Delta$. Given the uncertain nature of the HV behavior model, a probabilistic constraint, or chance constraint, is set to ensure a safe distance between the trailing AV and the HV. This constraint is expressed as:
\begin{equation} 
    \mathrm{Pr}\left(p_{k}^{N_a}-(p_{k}^{H} + \Delta) > \Delta_\text{ext} \right) \geq p_{\text{def}} \, . \tag{13} \label{eqn:chance_constraint} 
\end{equation}
Here, $\Delta$ denotes the predefined safe distance between AVs, and $\Delta_\text{ext} \geq 0$ is an extra distance established to account for the HV's stochastic behavior. The desired satisfaction probability is denoted as $p_{\text{def}}$. To rephrase the distance constraint $\mathcal{X}$, we can express it as a single half-space constraint $\mathcal{X}^{hs} := \bigl\{x \vert h^{\top}x \leq b \bigl\}$, $ h \in \mathbb{R}^n$, where $ h \in \mathbb{R}^n$, and $b \in \mathbb{R}$. As derived in \cite{hewing2019}, a method to tighten the constraint on the state mean is expressed as:
\begin{equation} 
    \mathcal{X}^{h s}\left(\Sigma_{i}^{x}\right):=\left\{x \mid h^{\top} x \leq b-\phi^{-1}\left(p_{\text{def}}\right) \sqrt{h^{\top} \Sigma_{i}^{x} h}\right\} \, . \tag{14} \label{eqn:18} 
\end{equation}
In this equation, $h^{\top} = \begin{bmatrix} -1 & 1\end{bmatrix}$, $x := \begin{bmatrix} p_{k}^{N_a} & p_{k}^{H}+\Delta \end{bmatrix}^{\top}$, and $b = -\Delta_\text{ext}$. Here, $\phi^{-1}$ is the inverse of the cumulative distribution function (CDF). In our scenario:
\begin{equation} 
    \Sigma_{k}^{x} := \begin{bmatrix} \Sigma_{k}^{p^{N_a}} \\ \Sigma_{k}^{p^{H}}+\Delta \end{bmatrix} = \begin{bmatrix} 0 & 0 \\ 0 & \Sigma_{k}^{p^{H}}\end{bmatrix} \, . \tag{15} \label{eqn:19} 
\end{equation}
Here, there is no covariance between the position of the HV and the AV leading it, indicated by $\Sigma_{k}^{p^{N_a}} = 0$. The position variance of the HV, denoted as $\Sigma_{k}^{p^{H}}$, is computed using \eqref{eqn:variance_prop_d}. To establish a more ``tightened'' constraint on the position state, we substitute \eqref{eqn:19} into \eqref{eqn:18} to obtain:
\begin{equation} 
    p_{k}^{N_a}-p_{k}^{H} \geq \Delta + \Delta_\text{ext} + \phi^{-1}\left(p_{\text{def}}\right) \sqrt{ \Sigma_{k}^{p^{H}}} \, . \tag{16} \label{eqn:safe_dis_HV} 
\end{equation}

The chance constraint for safe distance, given by \eqref{eqn:chance_constraint}, can be reduced to a deterministic equation as shown in \eqref{eqn:safe_dis_HV}. It can be further simplified by omitting the extra term $\Delta_\text{ext}$ with a high value definition for the satisfaction probability $p_{\text{def}}$. The safe distance between the AV and HV is adaptively adjusted, utilizing the estimated uncertainties from the HV's GP model as outlined in \eqref{eqn:variance_prop_d}. This adjustment ensures the safe distance always exceeds $\Delta$ under all conditions.

\subsection{GP-Based MPC}
\label{sec:GP-MPC}
We designed an MPC strategy integrating our proposed HV model, suitable for scenarios involving a mixed-vehicle platoon of $N_a$ AVs and HV, as illustrated in Fig. \ref{figure:mixed_platoon}. The implemented strategy can be represented as:
\begin{ceqn}
    \begin{align} 
       \underset{\mathbb{V}}{\text{min}} \sum_{\mathbf{n_a}=1}^{N_a} & \sum_{i=k}^{k+N-1} \Big\| \mathrm{acc}_{{i}|k}^{\mathbf{n_a}} \Big\|^2_R + \sum_{i=k}^{k+N} \Big\| v _{{i+1}|k}^1 - v_{{i+1}|k}^\text{ref} \Big\|^2_{Q_1} \nonumber \\
       & \qquad \quad + \sum_{\mathbf{n_a}=2}^{N_a}\sum_{i=k}^{k+N} \Big\| v_{{i+1}|k}^{\mathbf{n_a}} - v_{{i+1}|k}^{\mathbf{n_a}-1} \Big\|^2_{Q_2} \tag{17a} \label{eqn:mpc_a} \\
       \text{with} \ \mathbb{V} = & \left\{v_{{i}|k}^1, v_{{i}|k}^{\mathbf{n_a}}, v_{{i}|k}^{H}, p_{{i}|k}^{\mathbf{n_a}}, \mu_{{i}|k}^{p^{H}}, \Sigma_{{i}|k}^{p^{H}}, \mathrm{acc}_{{i}|k}^{\mathbf{n_a}} \right\} \, \nonumber \\
        \text {subject to} \nonumber \\
        v_{{i+1}|k}^{\mathbf{n_a}} &= v_{{i}|k}^{\mathbf{n_a}} + T \, \mathrm{acc}_{{i}|k}^{\mathbf{n_a}} , \  p_{{i+1}|k}^{\mathbf{n_a}} = p_{{i}|k}^{\mathbf{n_a}} + T \, v_{{i}|k}^{\mathbf{n_a}} \, , \nonumber \\
        \mathbf{n_a} &= \{1, 2, \cdots, N_a\}  \, , \tag{17b} \label{eqn:mpc_b} \\ 
        v^{H}_{{i}|k} &= {f}\left(v_{i-1:i-4 | k}^{H}, {v}_{i-1:i-4 | k}^{N_a} \right) \, , \tag{17c} \label{eqn:mpc_c}\\
        \mu_{{i+1}|k}^{p^{H}} &= \mu_{{i}|k}^{p^{H}} + T \, v_{{i}|k}^{H} + T \, \mu^d (v_{{i-1}|{k}}^{H}, {v}_{{i-1}|{k}}^{N_a}) \, , \tag{17d} \label{eqn:mpc_d}\\
        \Sigma_{{i+1}|k}^{p^{H}} &= \Sigma_{{i}|k}^{p^{H}} + T^2 \Sigma^d (v_{{i-1}|{k}}^{H}, {v}_{{i-1}|{k}}^{N_a}) \, , \tag{17e} \label{eqn:mpc_e}\\
        p_{{i}|k}^{\mathbf{n_a}-1} & - p_{{i}|k}^{\mathbf{n_a}} \geq \Delta \, , \tag{17f} \label{eqn:mpc_f}\\
        p_{{i}|k}^{N_a} & - \mu_{{i}|k}^{p^{H}} \geq \Delta + \phi^{-1}\left(p_{\text{def}}\right) \sqrt{ \Sigma_{{i}|k}^{p^{H}}} \, , \tag{17g} \label{eqn:mpc_g}\\
        v_{\text{min}} & \leq v_{{i}|k}^{\mathbf{n_a}} \leq v_{\text{max}} \, , \ \mathrm{acc}_{\text{min}} \leq \mathrm{acc}_{{i}|k}^{\mathbf{n_a}} \leq \mathrm{acc}_{\text{max}} \, . \tag{17h} \label{eqn:mpc_h}
    \end{align}
\end{ceqn}

Considering the uncontrollable nature of human drivers, the cost function \eqref{eqn:mpc_a} does not account for HV velocities. Instead, it primarily focuses on the divergence between the reference velocity ($v_{i}^\text{ref}$) and the leader AV's velocity ($v_{i}^1$), velocity differences between adjacent AVs ($v_{i}^{\mathbf{n_a}} - v_{i}^{\mathbf{n_a}-1}$), and control inputs ($\mathrm{acc}_{i}^{\mathbf{n_a}}$), weighted by positive constants $Q_1$, $Q_2$, and $R$. However, HV velocities are incorporated in the GP+ARX model of the HV, which takes into account the compensated HV velocities in \eqref{eqn:mean_prop_d} and propagates HV variances as described in \eqref{eqn:variance_prop_d}. An adaptive safe distance between AV and HV is integrated, utilizing the HV variance propagation from \eqref{eqn:safe_dis_HV}. These components are considered as equality and inequality constraints in the MPC policy formulation. It is important to note that all variable definitions in these equality and inequality constraint equations remain consistent with their original derivation in the earlier sections.

More specifically, the equality constraints \eqref{eqn:mpc_b}$-$\eqref{eqn:mpc_e} specify the AV model as defined in \eqref{eqn:av_eqn_a} and \eqref{eqn:av_eqn_b}, together with the HV's ARX model, represented by \eqref{eqn:arx}, as expressed in \eqref{eqn:mpc_c}. The HV mean and variance position propagation equations, originally derived in \eqref{eqn:mean_prop_d} and \eqref{eqn:variance_prop_d}, are represented by \eqref{eqn:mpc_d} and \eqref{eqn:mpc_e}. In these equations, $\mu^d(\cdot)$ and $\Sigma^d(\cdot)$ denote the mean and variance predictions of the GP model defined in \eqref{eqn:gp_eqn}. The inequality constraints include the velocity and acceleration conditions \eqref{eqn:mpc_h}, as well as the safe distance requirements \eqref{eqn:mpc_f} and \eqref{eqn:mpc_g} (originally derived in \eqref{eqn:safe_dis_HV}).

\subsection{Dynamic Sparse GP Prediction in MPC}
\label{sec:dynamic_sparseGP_mpc}
In Sec. \ref{sec:spares_gp_arx}, we showed the effectiveness of the sparse GP+ARX model in reducing computation time and enhancing modeling accuracy for HV velocity predictions compared to the standard GP+ARX and ARX models. Nonetheless, the real-time computation integration of GP models into the MPC framework remains a challenge due to the already computationally heavy nature of MPC, which involves resolving an optimal control problem at each time step while complying with the constraints specified in equations \eqref{eqn:mpc_b}--\eqref{eqn:mpc_h}. To tackle this issue, we implemented a dynamic sparse approximation of the GP-MPC method, proposed by \cite{hewing2019}, for additional speed enhancements.

Considering the receding horizon nature of MPC, the predictive trajectory at the current time step is similar to the one calculated at the preceding time step. Building upon this fact, we employed the sparse GP to perform calculations on the trajectory derived at the previous time step. These results were then used to propagate the mean and variance at the current time step by applying equations \eqref{eqn:mpc_d} and \eqref{eqn:mpc_e}. This approach alleviates the computational burden by replacing multiple predictions for each time step within the horizon with a single sparse GP prediction for the entire prediction horizon. Consequently, it makes the GP-MPC scheme feasible for real-time applications that demand fast sampling rates.

\subsection{Discussions}
\label{sec:discussions_control}

In the current GP-MPC strategy, we focus on improving safety in longitudinal car-following control for AV platoons in mixed traffic scenarios. However, recognizing the importance of overall traffic system efficiency, or system utility \cite{zhong2023hierarchical}, the integration of utility metrics into our control objectives represents a substantial opportunity for enhancing global traffic management. System utility, encompassing factors like traffic flow efficiency, average travel time, and fuel consumption \cite{qadri2020state}, can transform our safety-focused controller into a multifaceted traffic management tool. By adapting the control objective function of our GP-MPC to include these metrics, we could aim to optimize a more comprehensive traffic management strategy. This requires extending the current framework to a multi-objective optimization function, balancing vehicle platoon safety with broader traffic efficiency metrics. For example, adapting the MPC cost function to include terms penalizing deviations from desired traffic flow patterns while ensuring safety inside the mixed vehicle platoon. This approach aligns with recent studies indicating that varying CAV penetration rates and spatial distributions significantly influence traffic efficiency and safety \cite{gueriau2020quantifying, dong2022impact}. To achieve this, one significant challenge is formulating these additional objectives in a way that they can be integrated into the existing sparse GP model without substantially increasing computational complexity. 

Different from a notable recent work on AV platoons following HVs \cite{huang2020learning}, we primarily focused on HVs trailing AV platoons in mixed traffic scenarios. While this choice reflects a prevalent research motivation that a significant percentage of mixed traffic accidents involve HVs rear-ending AVs, as highlighted in Sec. \ref{sec:intro}, we acknowledge the complexities of HVs within AV platoons \cite{di2021survey}. Theoretically, our model is adaptable to such configurations by segmenting the traffic into two separate platoons: one AV platoon leading HVs and another AV platoon following HVs. While we have not explicitly tested our method for this setup, the similarities in the underlying dynamics and control principles suggest the feasibility of extending our approach. Future work will explore this aspect, enhancing the strategy to manage mixed platoons with varying vehicle sequences. 

The current scope of our work is to demonstrate the feasibility and effectiveness of the GP-based MPC utilizing our proposed HV model to enhance safety in a mixed-vehicle platoon, without conducting a formal analysis of stability. Despite the increasing interest in GP-based MPC, establishing stability is still an open challenge \cite{berberich2020}. There are currently no adaptive MPC frameworks, GP-based MPC included, that can ensure recursive feasibility and stability for nonlinear systems in state space under probabilistic constraints \cite{bujarbaruah2020} due to the complexity of GP models and the inherent intricacies of MPC formulations \cite{zhang2022}. Therefore, our strategy emphasizes demonstrating feasibility via simulations and applications in real-world systems, highlighting its ability to handle constraints. 

\section{Simulations}
\label{sec:simulations}

This section demonstrates the effectiveness of our sparse Gaussian process-based model predictive control (GP-MPC) strategy in managing mixed-traffic environments. We aim to assess the GP-MPC's advancements in enhancing safety, improving travel speeds, and optimizing computational efficiency. To achieve this, we conduct a series of simulation tests comparing the GP-MPC with a baseline standard MPC. These tests include emergency braking scenarios (Sec. \ref{sec:emergency}) and utilize the worldwide harmonized light vehicle test procedure (WLTP) \cite{coppola2022eco} as the reference speed for the leader AV (Sec. \ref{sec:wltp}). Additionally, we explore the practical implementation of our GP-MPC strategy in real-time operation settings, moving beyond theoretical analysis (Sec. \ref{sec:real-time}).

For this section, we have not included the simulation results using the standard GP model in HV modeling, as detailed in our preliminary work \cite{wang2024improving}. The exclusion is due to the impractical computation time required by the standard GP+ARX model for HV modeling, which demanded approximately 19 seconds per time step within the MPC loop. Such extensive computation time makes the standard GP model unsuitable for real-world applications, focusing our current study on more practical and efficient alternatives that offer real-time applicability and relevance in mixed-traffic scenarios.

\subsection{Setups}
\label{sec:setups}

\textbf{Baseline MPC:} As a comparative benchmark, we employed a standard MPC, hereafter referred to as the nominal MPC. This controller utilizes the ARX model of the HV for prediction purposes, as specified in equations \eqref{eqn:mpc_c} and \eqref{eqn:mpc_d} but excluding the third GP component. The distance constraint between the HV and the second AV, given by $\Delta$, does not incorporate the second adaptive component in equation \eqref{eqn:mpc_g}. Moreover, the nominal MPC does not account for the position variance propagation defined in equation \eqref{eqn:mpc_e}. However, it is important to note that while the control policies of the nominal MPC and our sparse GP-based MPC differ, both employ the same ARX+GP model as derived in equations \eqref{eqn:sys_model_a} and \eqref{eqn:sys_model_b} for simulating the HV model, ensuring a consistent simulation and enabling a direct comparison solely on the control policies.

\textbf{Simulation Environment:} The simulations were conducted using MATLAB R2022a on an Ubuntu 20.04 laptop.
Our simulation code is available at: \url{https://github.com/jwangjie/Mixed-Traffic-GP-MPC}. It is worth noting that even though all of our test cases in this section included two AVs in the AV platoon, our algorithms and code are scalable to manage larger platoon sizes.

\textbf{Initial Conditions and Constraints:} All vehicles in the simulations started with zero velocity to standardize initial conditions. The minimum safe distance between AVs and the deterministic component of AV-HV distance was set to $\Delta =$ 10\,m (equations \eqref{eqn:mpc_f} and \eqref{eqn:mpc_g}). The first leader AV was positioned at $p = 0$, the second AV was placed $1.2\,\Delta=$\,12\,m behind the first AV, and the HV was set $1.2\,\Delta=$\,12\,m behind the second AV. This configuration was chosen to ensure the desired spacing within the platoon at the beginning of the simulations. The chance constraint satisfaction probability in \eqref{eqn:mpc_g} was set at $p_{\text{def}}=0.95$. We defined the maximum and minimum accelerations as $a_{\text{max}}=$\,4\,m$^2/$s and $a_{\text{min}}=$\,$-4$\,m$^2/$s, respectively. The maximum and minimum velocities were set as $v_{\text{max}}=$\,37\,m$/$s (as the maximum speed in WLTP is 36.5\,m$/$s) and $v_{\text{min}}=$\,0\,m$/$s, respectively. 

\textbf{Sample Time:} In our simulations for emergency braking (Sec. \ref{sec:emergency}) and WLTP (Sec. \ref{sec:wltp}), we set the sample time for both GP-MPC and nominal MPC at $T =$ 0.1\,s. It's important to note that, due to computational constraints (using a laptop) and the inefficiency of MATLAB, each optimization in the MPC loops for both controllers requires approximately 0.2\,s per time step. This necessitates a step-by-step operation instead of real-time simulation when using a smaller sample time like $T =$ 0.1\,s. Despite this, simulations still output results every 0.1 seconds. For assessing the real-time efficacy of the GP-MPC, in Sec. \ref{sec:real-time}, we increase the sample time to $T=$ 0.25\,s. This adjustment ensures the completion of each optimization step in both MPC loops.

\textbf{Cost Function Weights:} To set the $Q_1$, $Q_2$, and $R$ weights of our cost function \eqref{eqn:mpc_a}, we applied a systematic method described in \cite{coppola2022eco}. Specifically tailored to our function, the control input weight ($R$) was set to double that of the reference speed tracking weights ($Q_1$ and $Q_2$). These parameters were then fine-tuned empirically. For emergency braking (Sec. \ref{sec:emergency}) and WLTP (Sec. \ref{sec:wltp}) scenarios, the final weights settled at $Q_1 = Q_2 = 5$ and $R = 10$. This convergence of initial estimations from the systematic method and the final weights after empirical fine-tuning underscores the robustness of our chosen weights and also highlights the reliability of the systematic approach in predicting effective weight values for our application. In contrast, for the real-time tests (Sec. \ref{sec:real-time}), we adjusted the weights to $Q_1 = Q_2 = 5$ and $R = 15$. This change was mainly to ensure safe distance constraints between AVs under larger sample times. 

\subsection{Emergency Braking Scenarios}
\label{sec:emergency}

In this section, we explore the performance of the nominal MPC and our proposed GP-MPC in emergency braking scenarios. These scenarios involve the leader AV in the platoon undergoing a series of abrupt speed reductions at predetermined times, simulating emergency braking conditions. The reference speed ($v^\text{ref}$) for the leader AV decreases from an initial 35\,m$/$s to 20\,m$/$s at 40\,s, 10\,m$/$s at 80\,s, 2\,m$/$s at 100\,s, and finally halts at 120\,s until the end of the simulation at 130\,s. The primary aim of these scenarios is to rigorously test the robustness and effectiveness of our HV modeling approach and the GP-MPC policy, especially in challenging conditions that demand quick adaptation to prevent rear-end collisions in AV-HV platooning.

\begin{figure}
    \centering
    \vspace{0.15cm}
    \subfloat{{\includegraphics[trim=0cm 0cm 0cm 0cm, width=0.98\columnwidth]{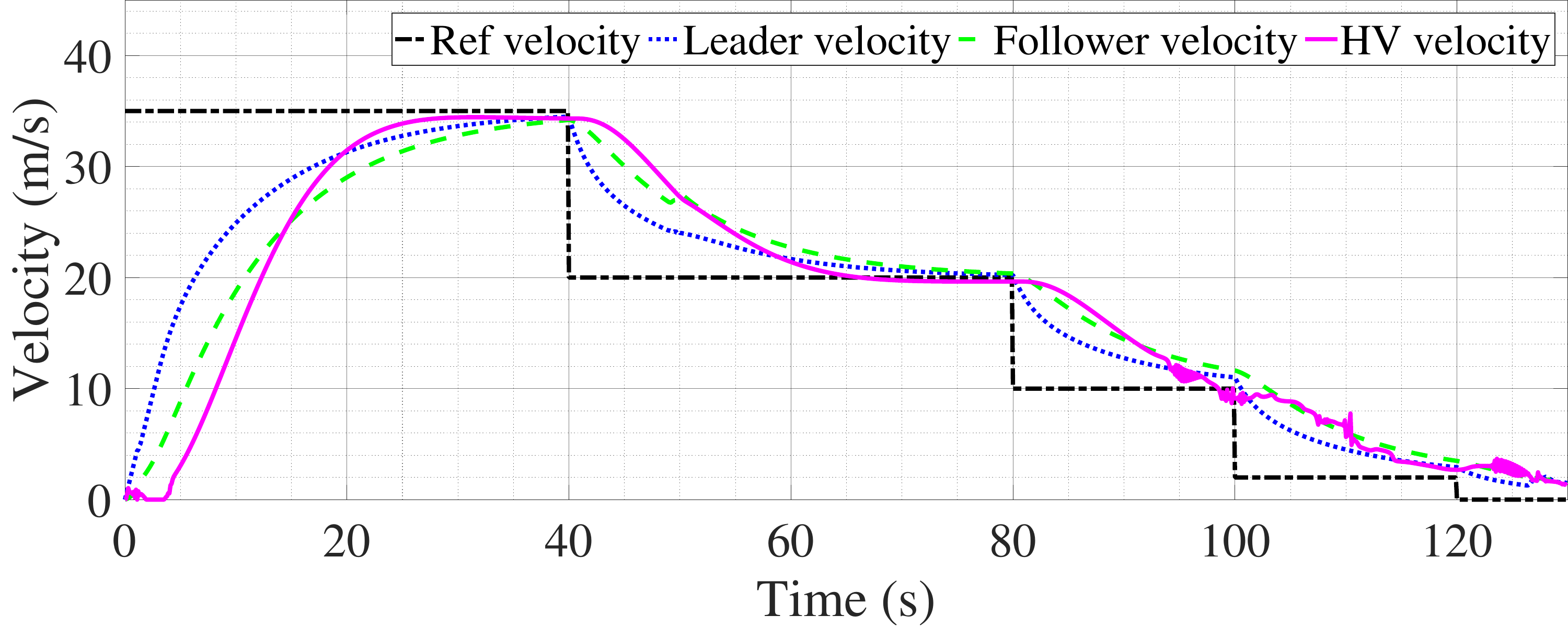} }}
    \qquad \qquad 
    \subfloat{{\includegraphics[trim=0cm 0cm 0cm 0.0cm, width=0.98\columnwidth]{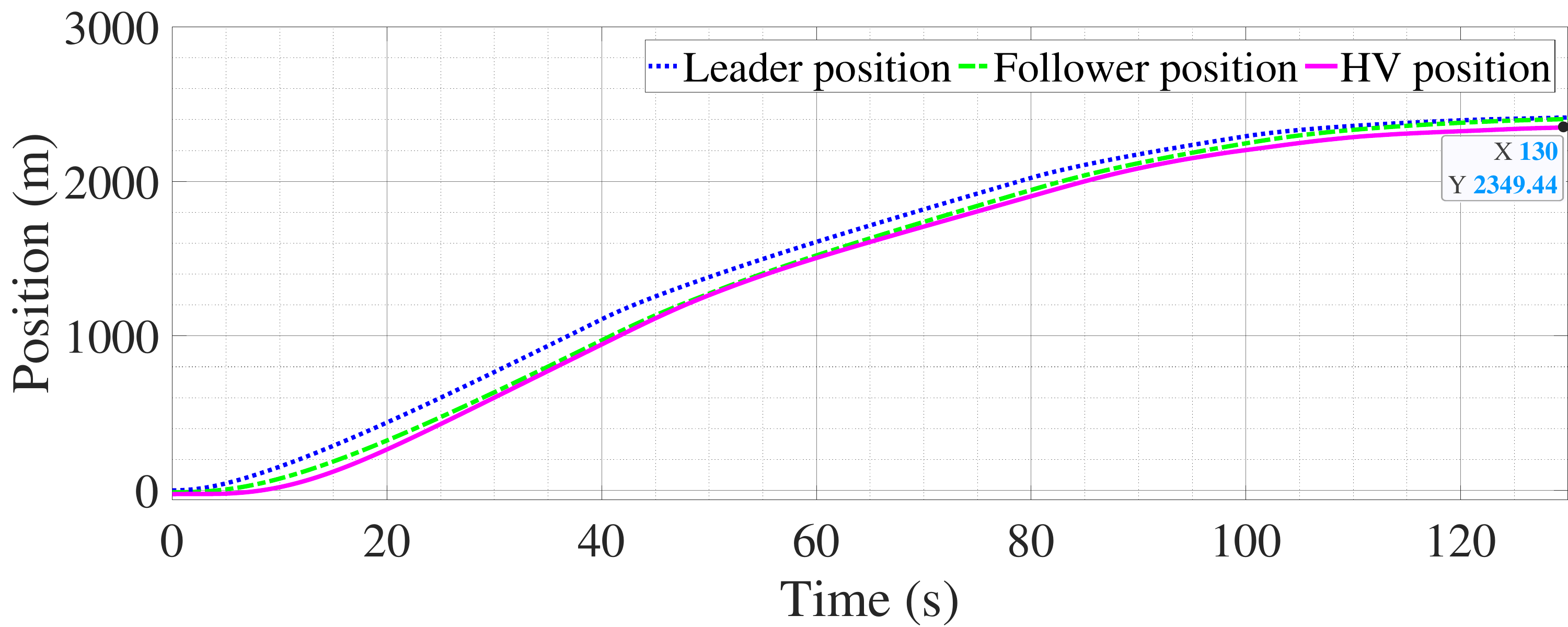} }}
    \qquad \qquad 
    \subfloat{{\includegraphics[trim=0.0cm 0cm 0.0cm 0.0cm, width=0.98\columnwidth]{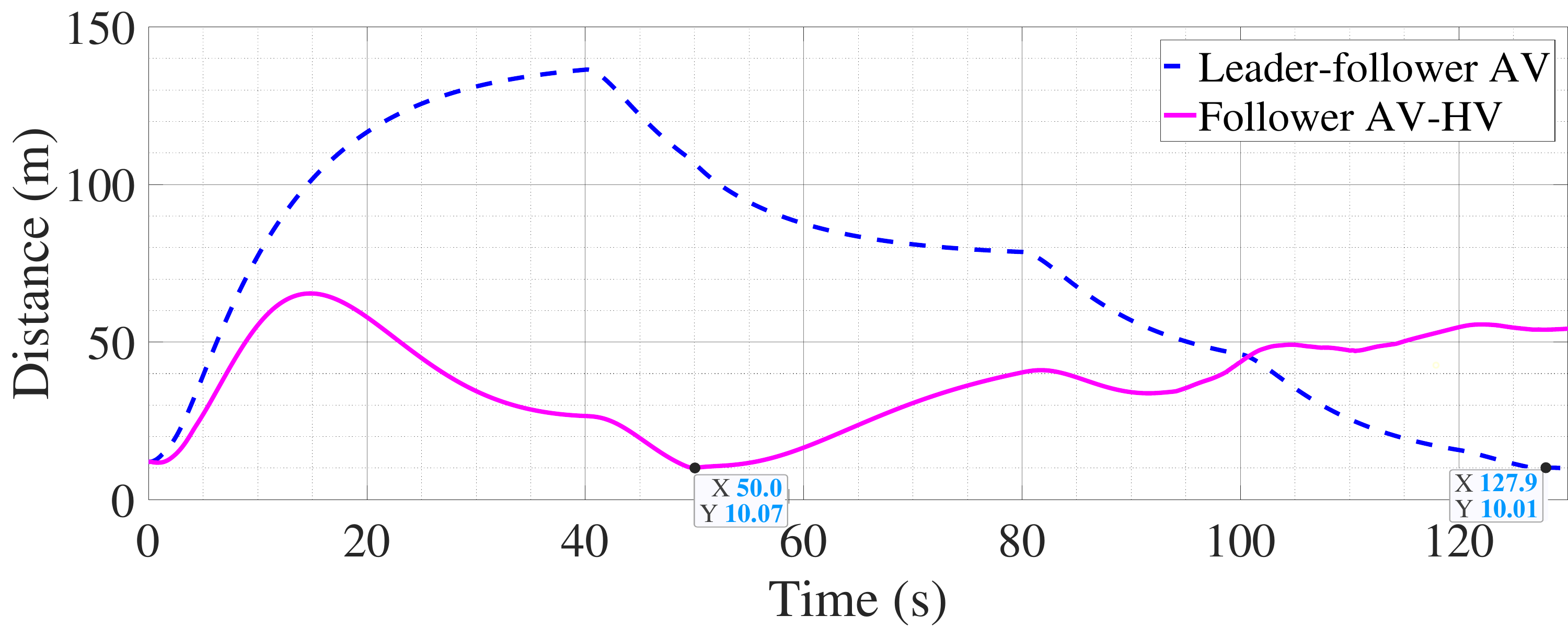} }}
    \caption{The simulation results from emergency braking scenarios using the nominal MPC. Arranged in a top-to-bottom sequence, the graphs depict the following: at the top, the velocity response of vehicles; in the middle, their position trajectories; and at the bottom, the inter-vehicle distances. These plots collectively illustrate how vehicles adapt their speed and positioning in response to emergency braking scenarios.
    }
\label{figure:nominal_simulation_braking}
\end{figure}

\begin{figure}
    \centering
    \vspace{0.15cm}
    \subfloat{{\includegraphics[trim=0cm 0cm 0cm 0cm, width=0.98\columnwidth]{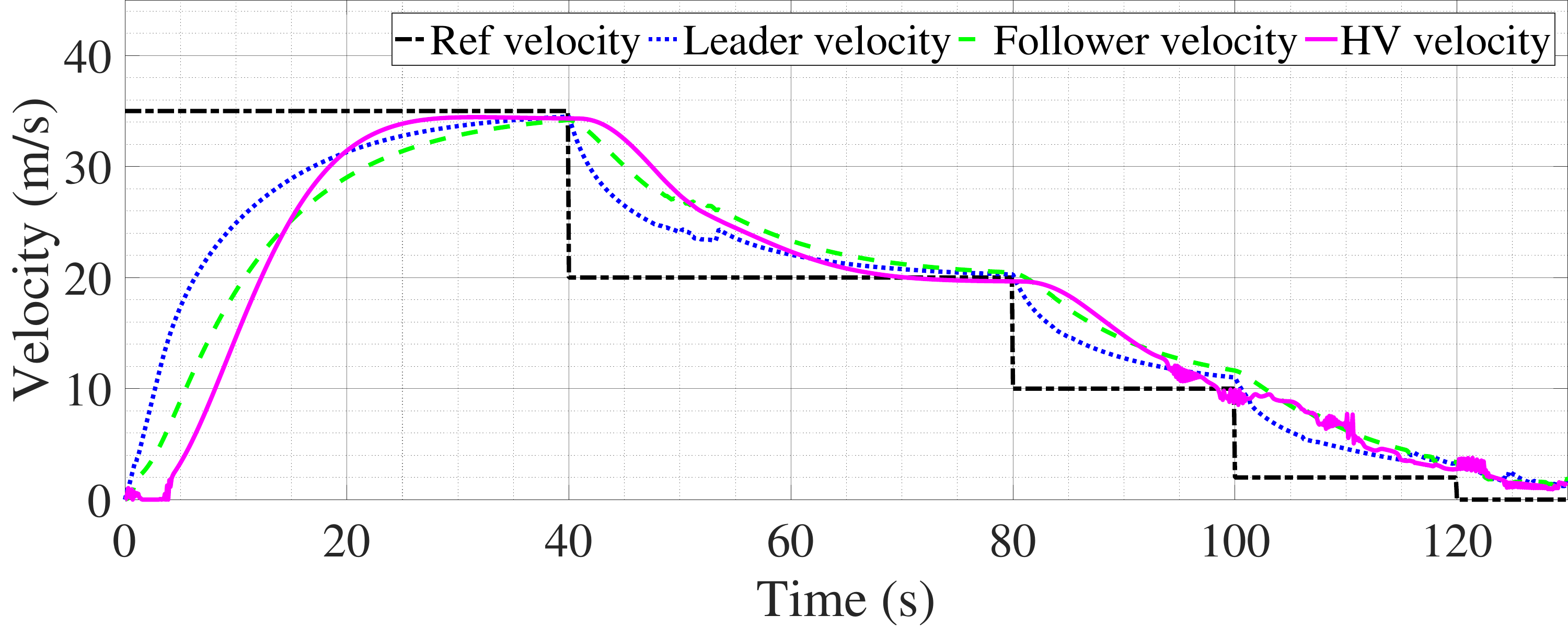} }}
    \qquad \qquad 
    \subfloat{{\includegraphics[trim=0cm 0cm 0cm 0.0cm, width=0.98\columnwidth]{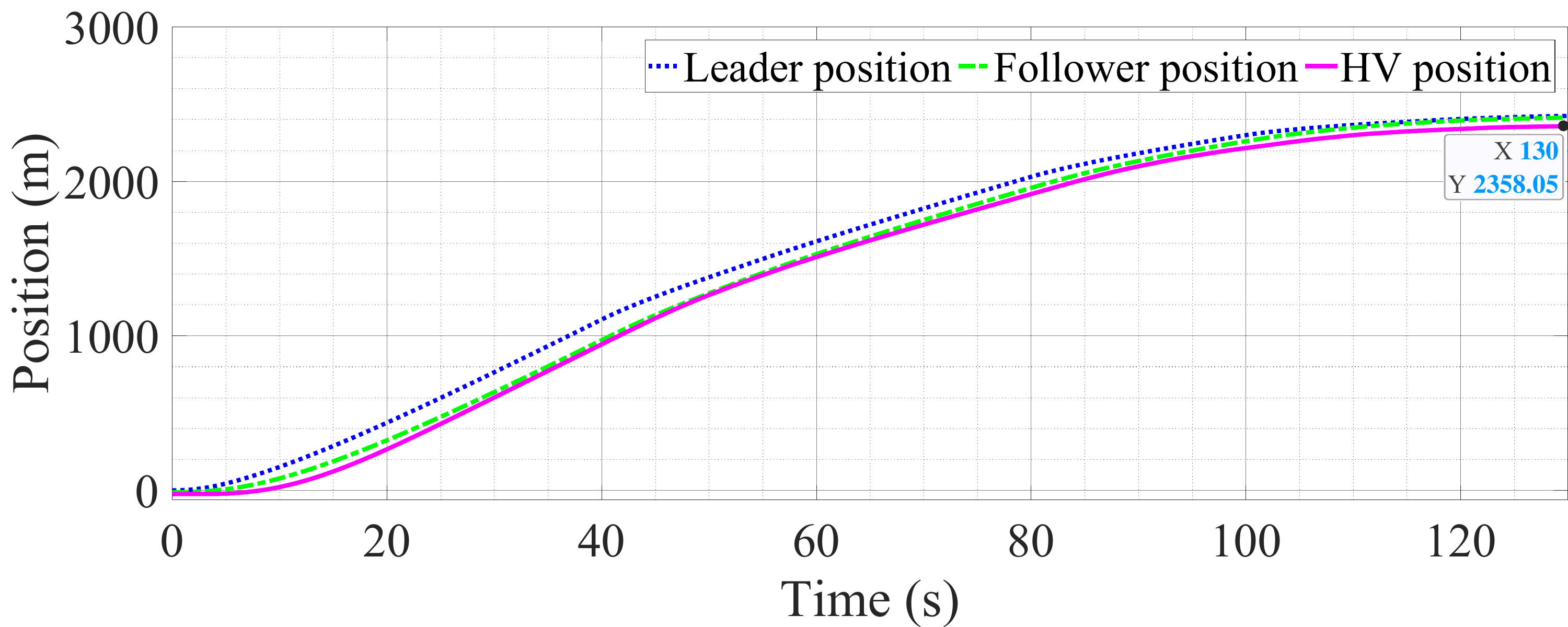} }}
    \qquad \qquad 
    \subfloat{{\includegraphics[trim=0.0cm 0cm 0.0cm 0.0cm, width=0.98\columnwidth]{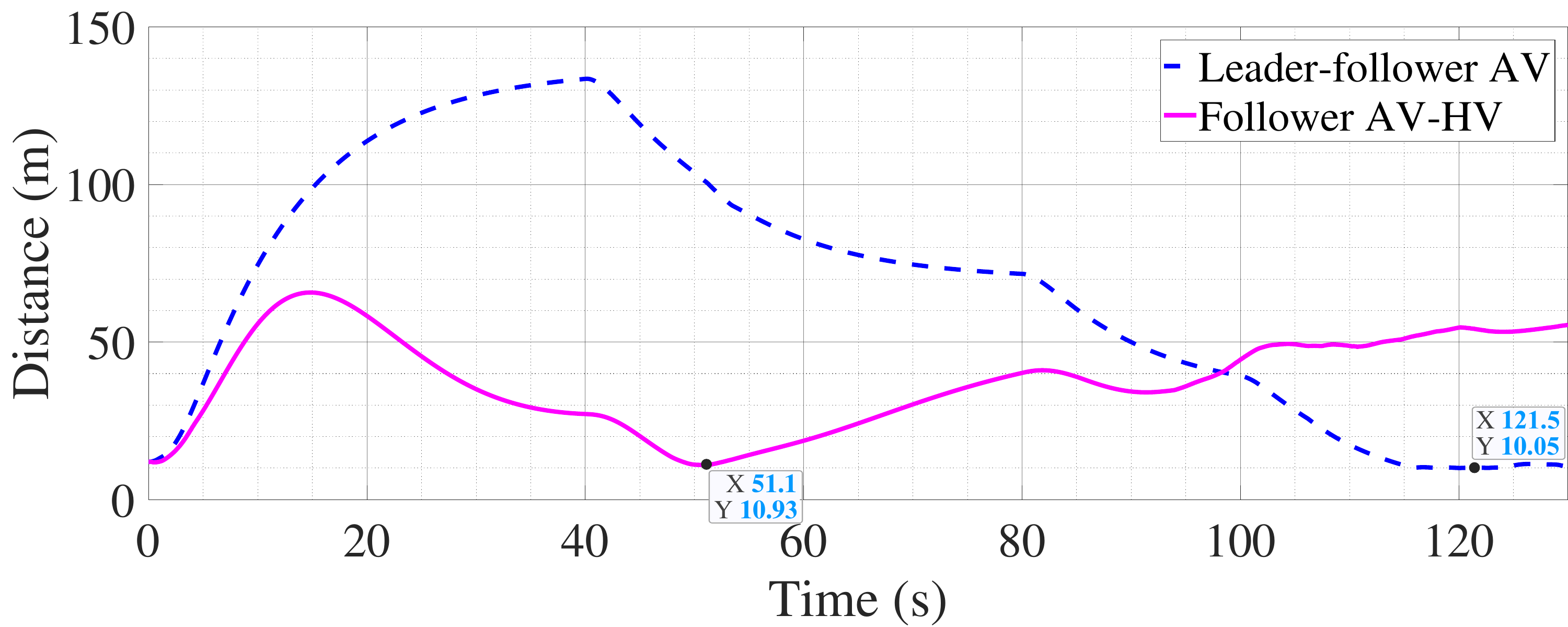} }}
    \caption{The simulation results from emergency braking scenarios using the GP-MPC. These plots, from top to bottom, represent velocity response, the position trajectories, and the inter-vehicle distance. Compared to the nominal MPC results in Fig. \ref{figure:nominal_simulation_braking}, the GP-MPC affords a 0.9-meter increase in the minimum distance between the HV and the trailing AV. This enhancement, alongside the extended distances covered by all vehicles under GP-MPC as detailed in Tab. \ref{tab:simulation_metrics_emergency}, indicates the model's superior capability in enhancing safety without imposing stricter control constraints.}
    \label{figure:gp_simulation_braking}
\end{figure}

To effectively visualize and analyze the results, we plotted velocity response, position trajectories, and inter-vehicle distance graphs in Fig. \ref{figure:nominal_simulation_braking} and \ref{figure:gp_simulation_braking} for the nominal MPC and GP-MPC, respectively. These plots show how vehicles in the platoon adjust their speeds in response to the changing reference speed during emergency braking. Both figures reveal some irregularities in HV behavior at some time stamps around 95\,s, 100\,s, 110\,s, and 125\,s. These irregularities are likely due to limitations in the training data for these specific velocity profiles, which may have impacted the precision of the GP model's predictions. While our current field data has effectively supported the calibration of the GP-based HV model in our model tests (as detailed in Sec. \ref{sec:spares_gp_arx}), it also highlights the need to capture the broader variability and complexity of human driving behavior. Future enhancements, including the enrichment of training data and refinement of the model, will focus on addressing these aspects. 

Additionally, there is a noticeable acceleration in the leader AVs starting at approximately $t=$ 126--128\,s and $t=$ 124--125\,s in Fig. \ref{figure:nominal_simulation_braking} and Fig. \ref{figure:gp_simulation_braking}, respectively. Examining this alongside the bottom inter-vehicle distance plots, it is evident that these accelerations aim to uphold the safe distance constraints as the distances between the leader-follower AVs approach the predefined minimum safe limit. This scenario exemplifies the distinction between instances where an HV interacts with a singular AV that lacks forward traffic cooperation and instances where AVs within a platoon coordinate their movements. Such synchronized movement in a platoon is pivotal for ensuring smoother and safer interactions between the trailing HV and the last AV, thus enhancing overall safety in mixed-traffic scenarios.

In Tab. \ref{tab:simulation_metrics_emergency}, we provided a numerical representation of the performance enhancements of the proposed GP-MPC over the nominal MPC (Nom. MPC in Tab. \ref{tab:simulation_metrics_emergency}). The GP-MPC obtained a minimum AV-HV distance of 10.93\,m, about 0.9\,m increase than the nominal MPC. This improvement is attributable to the inclusion of a GP uncertainty assessment component into the safe distance constraint, as detailed in \eqref{eqn:safe_dis_HV}. Extra adaptive components were added to the distance constraint in \eqref{eqn:mpc_g} at each time step in the GP-MPC, which led to more adaptive and responsive control. Furthermore, as shown in Tab. \ref{tab:simulation_metrics_emergency}, all vehicles in the GP-MPC model traveled approximately 10\,m more than vehicles in the nominal MPC scenario, indicating there is a more efficient traffic flow in the GP-MPC by enabling a higher overall speed for each vehicle in the mixed platoon.

\begin{table}
    \caption{Emergency Braking performance: The traveled distance of each vehicle and the minimum relative distance of AV-HV.
    }
    \label{tab:simulation_metrics_emergency}
    \begin{center}
        \begin{tabular}{|c|c|c|c|c|}
            \hline Controller & AV 1 & AV 2 & HV & Min AV-HV  \\
            \hline Nom. MPC & 2413.70\,m & 2403.69\,m & 2349.44\,m & 10.07\,m \\
            \hline GP-MPC & \textbf{2423.99}\,m & \textbf{2413.49}\,m &  \textbf{2358.05}\,m & \textbf{10.93}\,m\\
            \hline
        \end{tabular}
    \end{center}
\end{table}

\subsection{WLTP Scenario}
\label{sec:wltp}

\begin{figure}
    \centering
    \vspace{0.15cm}
    \subfloat{{\includegraphics[trim=0cm 0cm 0cm 0cm, width=0.98\columnwidth]{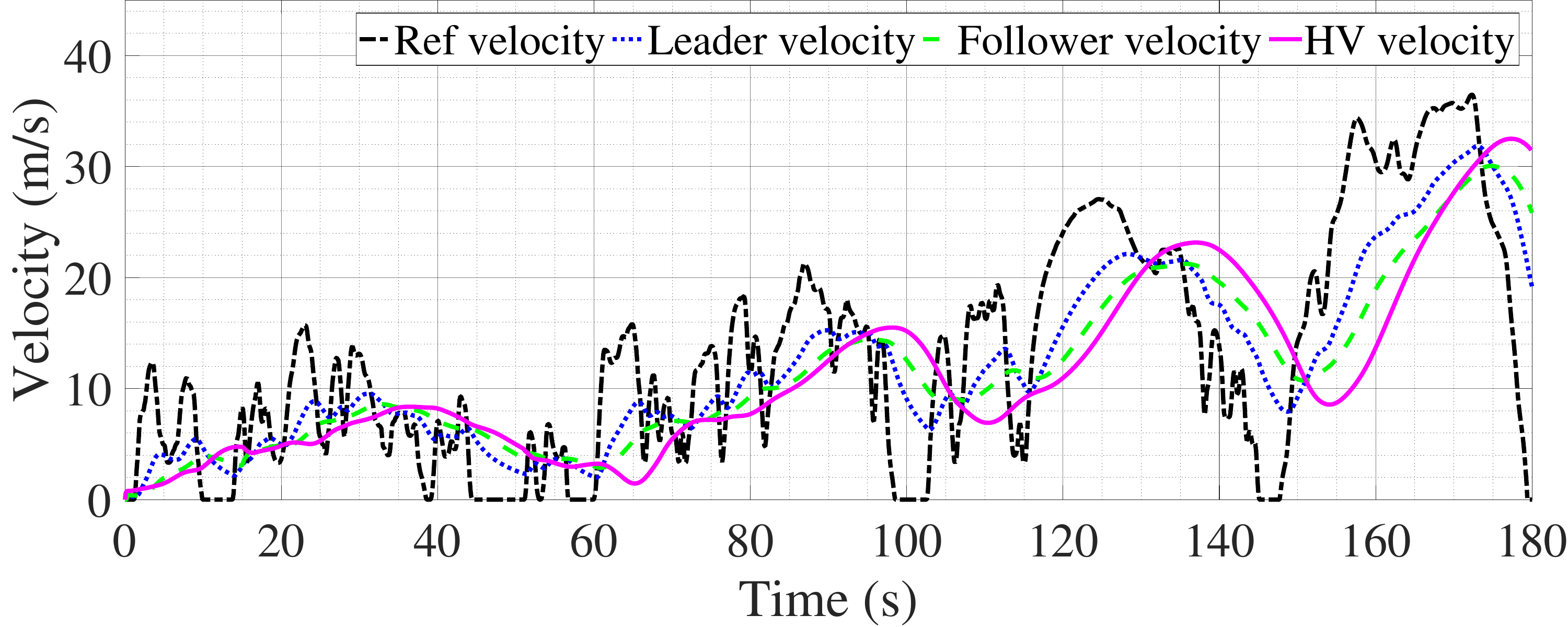} }}
    \qquad \qquad 
    \vspace{0.01cm}
    \subfloat{{\includegraphics[trim=0cm 0cm 0cm 0.0cm, width=0.98\columnwidth]{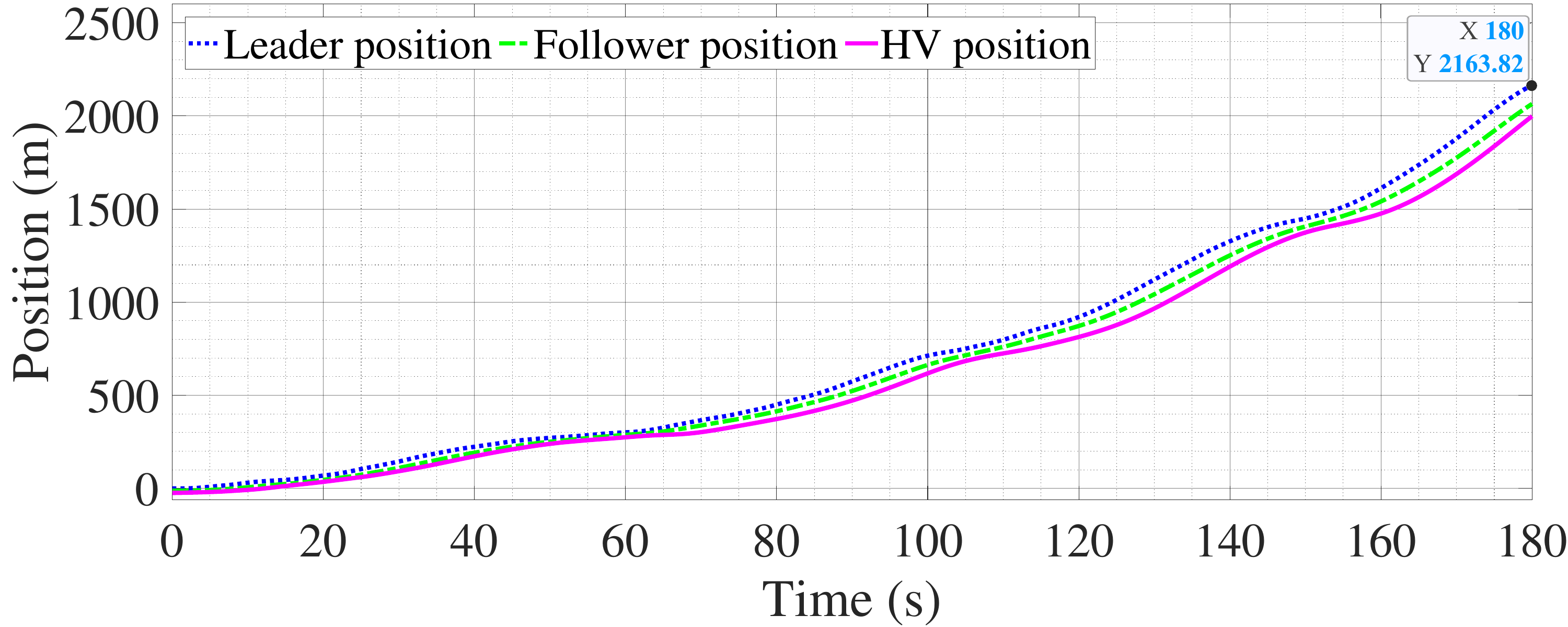} }}
    \qquad \qquad 
    \subfloat{{\includegraphics[trim=0.0cm 0cm 0.0cm 0.0cm, width=0.98\columnwidth]{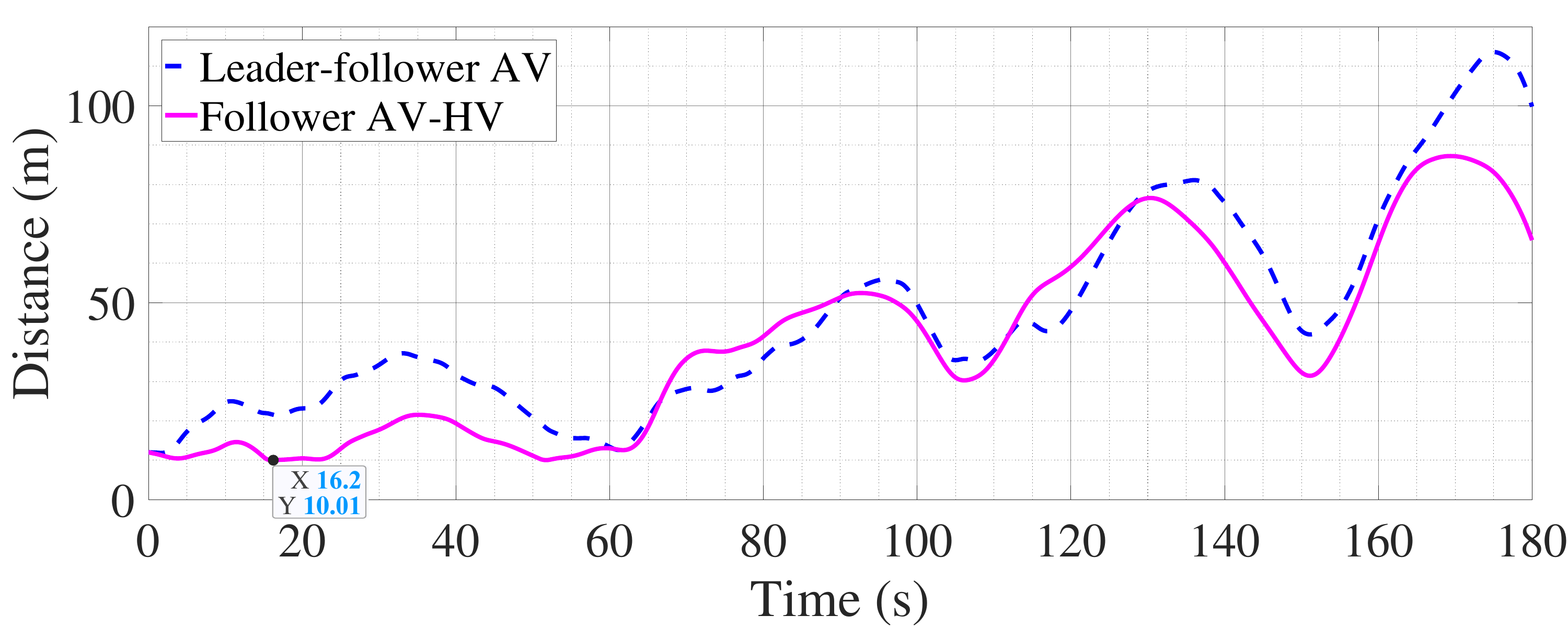} }}
    \caption{The simulation results of the WLTP scenario using the nominal MPC. The sequence of figures, shown from top to bottom, represents the velocity response, the position trajectories, and the inter-vehicle distances. These graphs provide a comprehensive overview of the adaptation to the WLTP's varied driving conditions, offering insights into the speed and spacing behaviors under this realistic testing protocol.
    }
\label{figure:nominal_simulation_wltp}
\end{figure}

\begin{figure}
    \centering
    \vspace{0.15cm}
    \subfloat{{\includegraphics[trim=0cm 0cm 0cm 0cm, width=0.98\columnwidth]{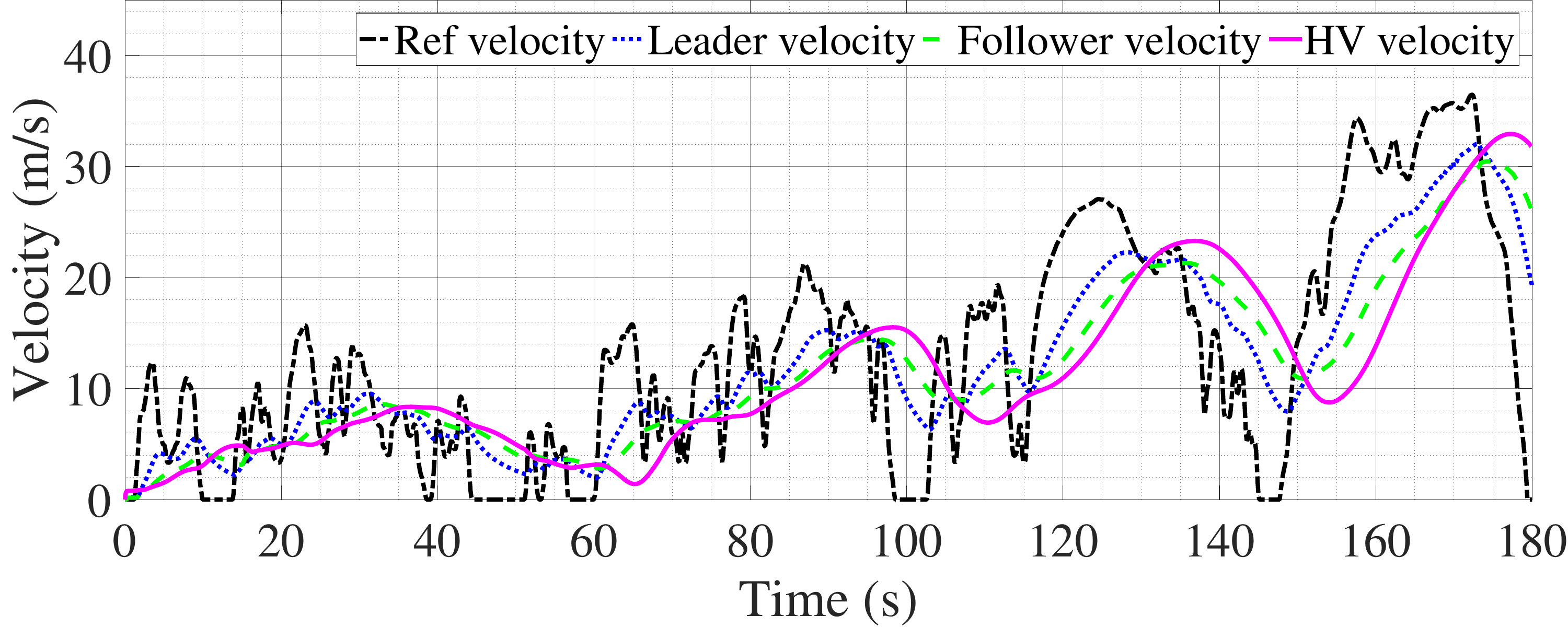} }}
    \qquad \qquad 
    \vspace{0.01cm}
    \subfloat{{\includegraphics[trim=0cm 0cm 0cm 0.0cm, width=0.98\columnwidth]{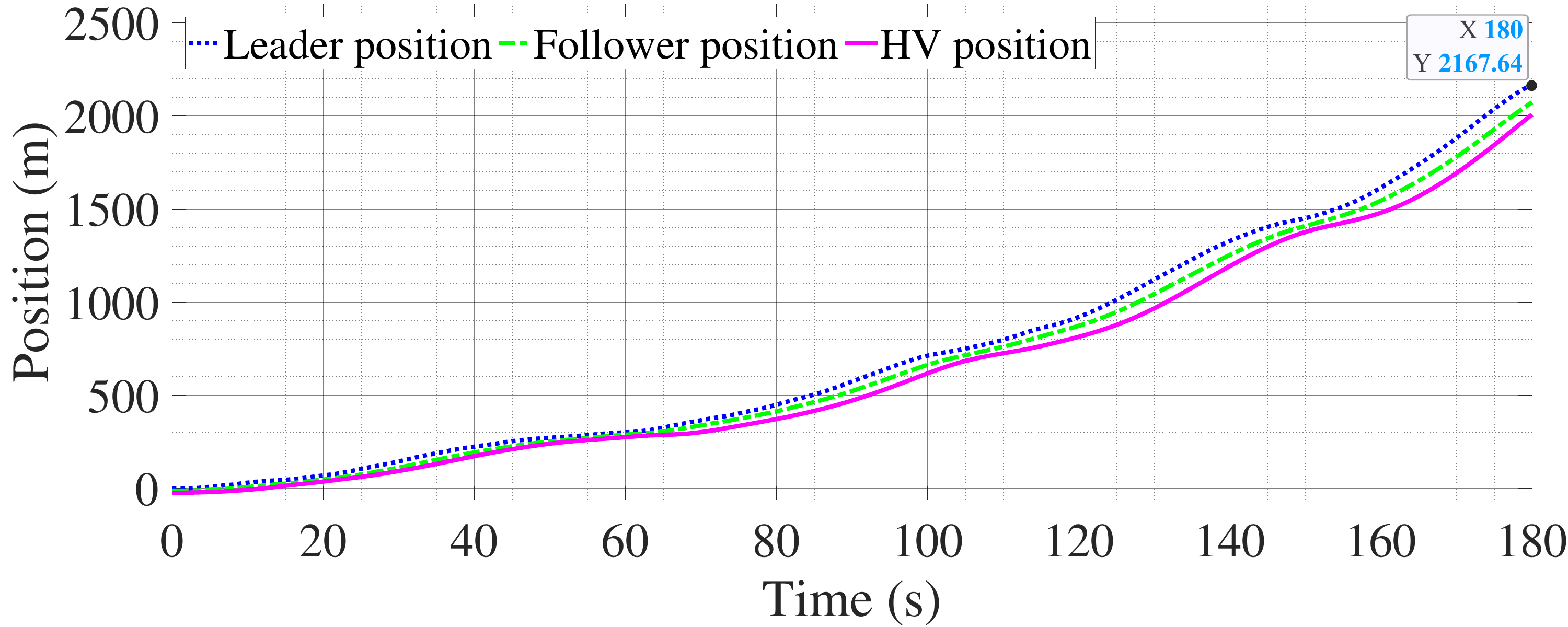} }}
    \qquad \qquad 
    \subfloat{{\includegraphics[trim=0.0cm 0cm 0.0cm 0.0cm, width=0.98\columnwidth]{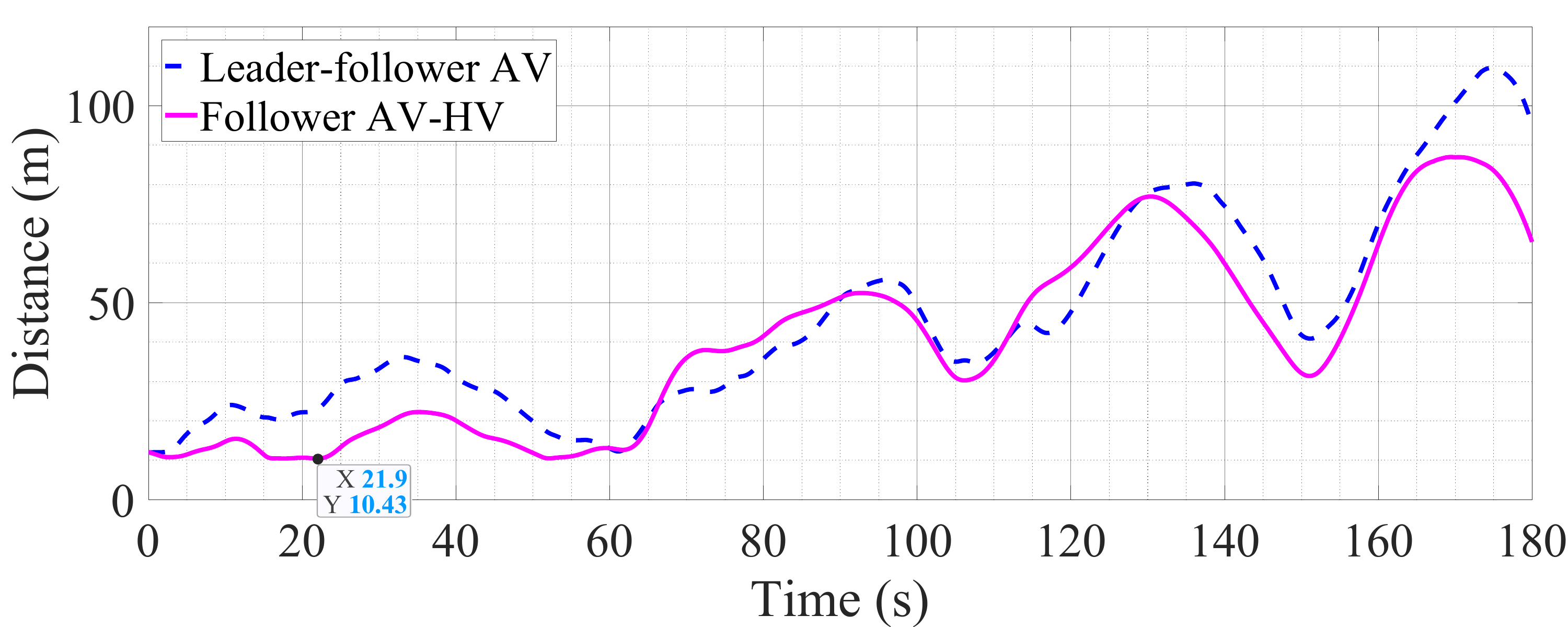} }}
    \caption{The simulation outcomes of the WLTP scenario using the GP-MPC. Arranged similarly in descending order, these plots encompass the velocity response, the position trajectories, and the inter-vehicle distances. Notably, when compared to the nominal MPC depicted in Fig. \ref{figure:nominal_simulation_wltp}, the GP-MPC demonstrates enhanced safety, increasing the minimum distance between the HV and trailing AV by 0.4 meters. As detailed in Tab. \ref{tab:simulation_metrics_wltp}, this model ensures all vehicles maintain greater distances, highlighting the GP-MPC's ability to improve safety and efficiency without additional operational constraints under this realistic traffic scenario.
    }
\label{figure:gp_simulation_wltp}
\end{figure}

To further assess the effectiveness of our GP-MPC policy, we conducted simulations using the worldwide harmonized light vehicle test procedure (WLTP) as the reference speed profile for the leader HV. Initially designed for evaluating pollutant and CO2 emissions and fuel consumption, the WLTP has also been established as a realistic driving profile for testing controller performance in car-following scenarios \cite{coppola2022eco}. It includes diverse driving phases including stops, accelerations, and braking, divided into four parts with varying average speeds: low, medium, high, and extra high. This comprehensive range of driving behaviors makes the WLTP an ideal benchmark for evaluating our controller's performance in more realistic car-following scenarios. The WLTP reference velocity profile, depicted in Fig. \ref{figure:nominal_simulation_wltp} and \ref{figure:gp_simulation_wltp} as dotted black lines, was adapted from data in \cite{zhu2023vehicle}, tailored to our specific sample time of $T =$ 0.1\,s over a 180\,s simulation duration.


We present the simulation results under the WLTP scenario using the nominal MPC and the GP-MPC in Fig. \ref{figure:nominal_simulation_wltp} and \ref{figure:gp_simulation_wltp}, respectively. These figures methodically illustrate the velocity response, position trajectories, and inter-vehicle distances of the vehicles, providing a detailed perspective on how each model reacts to the WLTP driving cycle's diverse conditions.

A notable observation is the smoother transition in vehicle speeds within these scenarios compared to emergency braking situations. This smoothness results from the WLTP's gradual changes in reference velocity. In terms of safety, the GP-MPC shows an increase in the minimum AV-HV distance to 10.43\,m, an improvement of about 0.4\,m over the nominal MPC. Although this increment is modest compared to the more dynamic emergency braking scenario, it aptly reflects more stable HV behaviors (with less uncertainty) in WLTP conditions. In these less dynamic scenarios, the reduced uncertainty assessments by our GP-based HV model led to a smaller addition to the adaptive safe distance constraints in \eqref{eqn:mpc_g}. This result highlights the method's capacity to adapt its safety margins according to the variability of driving conditions, thereby proving its effectiveness in different traffic scenarios.

Furthermore, as detailed in Tab. \ref{tab:simulation_metrics_wltp}, the GP-MPC not only improves safety margins but also enables higher overall speeds for vehicles. This is evident from the positions of all vehicles in the GP-MPC being several meters (4--9\,m) ahead of those in the nominal MPC, indicating a more efficient traffic flow under realistic driving scenarios.

\begin{table}
    \caption{WLTP scenario performance: The traveled distance of each vehicle and the minimum relative distance of AV-HV.
    }
    \label{tab:simulation_metrics_wltp}
    \begin{center}
        \begin{tabular}{|c|c|c|c|c|}
            \hline Controller & AV 1 & AV 2 & HV & Min AV-HV  \\
            \hline Nom. MPC & 2163.82\,m & 2064.06\,m & 1988.20\,m & 10.01\,m \\
            \hline GP-MPC & \textbf{2167.64}\,m & \textbf{2072.58}\,m &  \textbf{2007.16}\,m & \textbf{10.43}\,m\\
            \hline
        \end{tabular}
    \end{center}
\end{table}

\subsection{Real-Time Tests}
\label{sec:real-time}
This section extends our evaluation of the GP-MPC policy to a real-time operational scenario. Unlike the controlled conditions of the step-by-step simulations in emergency braking (Sec. \ref{sec:emergency}) and WLTP (Sec. \ref{sec:wltp}), by adjusting the sample time to $T =$ 0.25\,s, we ensure the completion of optimization calculations within each MPC loop's time step, thus enable conditions that more closely mimic real-world implementations. 

\begin{figure}
    \centering
    \vspace{0.15cm}
    \subfloat{{\includegraphics[trim=0cm 0cm 0cm 0cm, width=0.98\columnwidth]{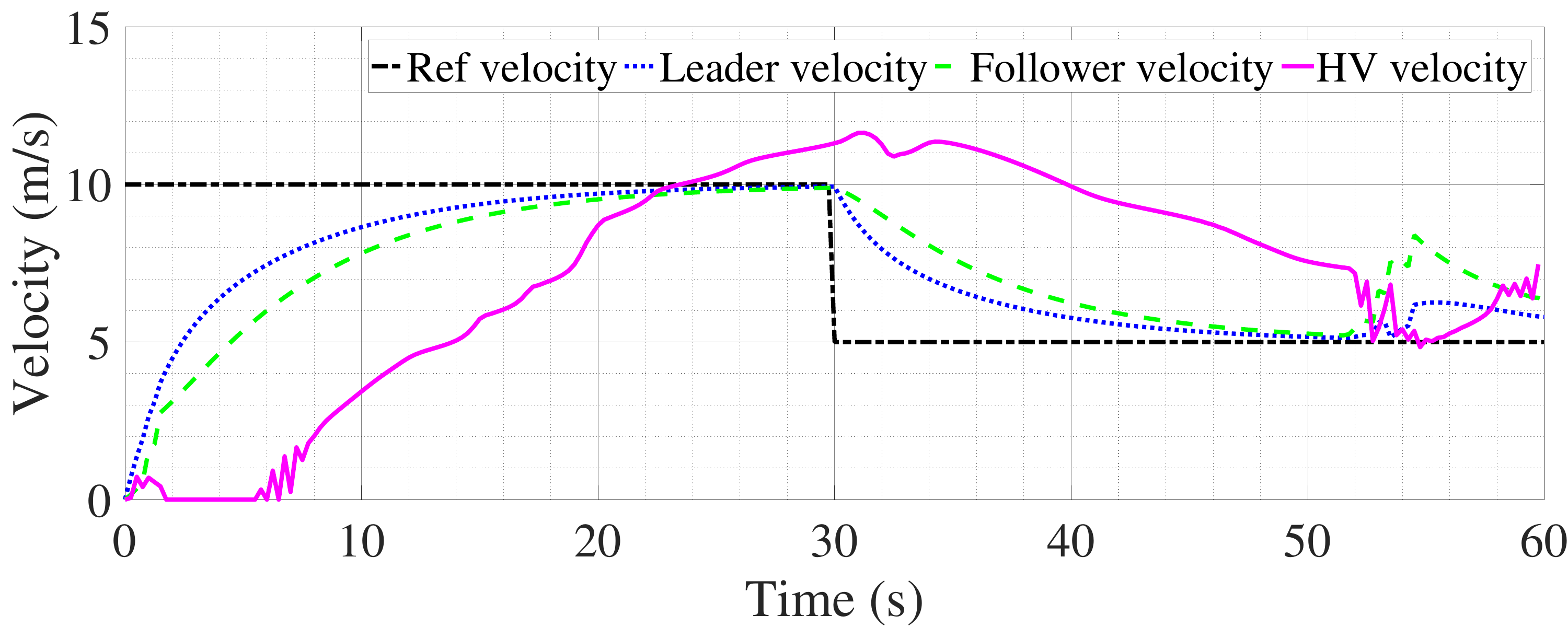} }}
    \qquad \qquad 
    \subfloat{{\includegraphics[trim=0cm 0cm 0cm 0.0cm, width=0.98\columnwidth]{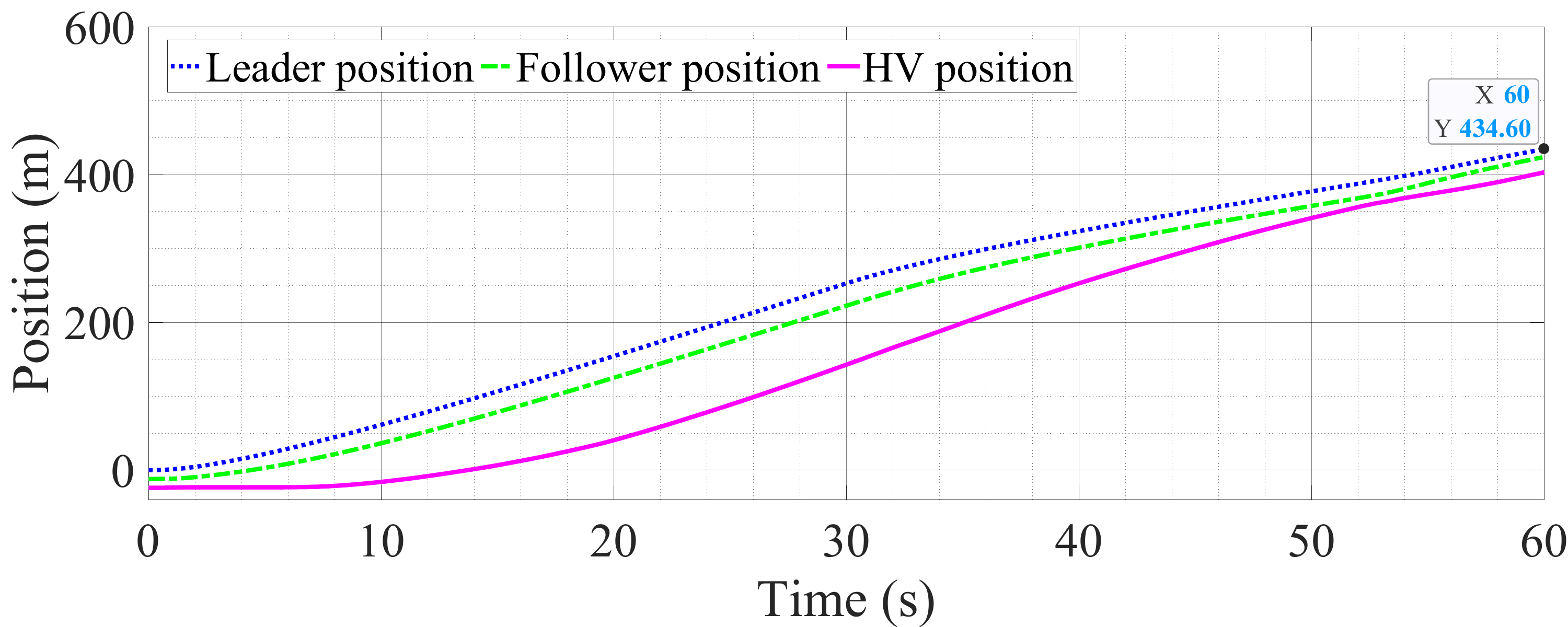} }}
    \qquad \qquad 
    \subfloat{{\includegraphics[trim=0.0cm 0cm 0.0cm 0.0cm, width=0.98\columnwidth]{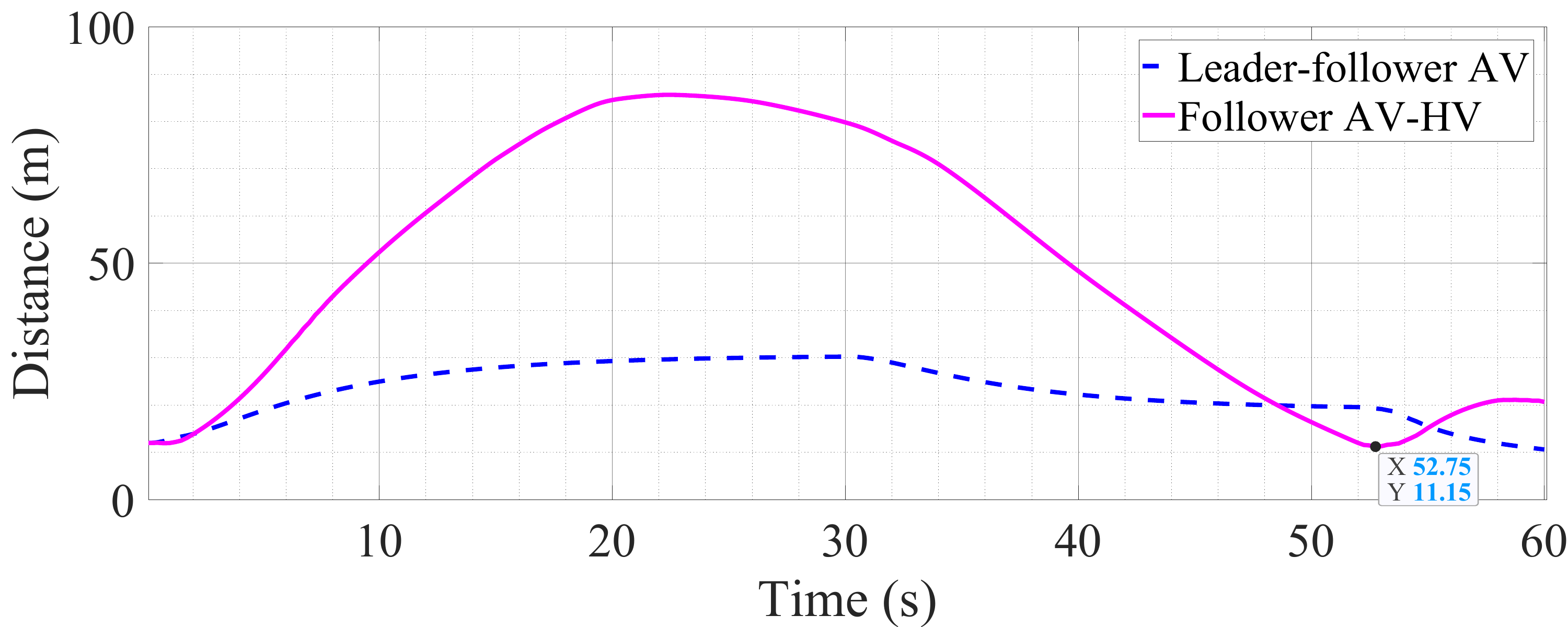} }}
    \caption{The simulation results of a low-speed emergency braking scenario using the nominal MPC in a real-time setting. The series of figures, organized from top to bottom, illustrate the velocity response, position trajectories, and inter-vehicle distances, providing insights into the controller's performance in managing vehicle dynamics and maintaining safety distances.
    }
\label{figure:nominal_simulation_realTime}
\end{figure}

\begin{figure}
    \centering
    \vspace{0.15cm}
    \subfloat{{\includegraphics[trim=0cm 0cm 0cm 0cm, width=0.98\columnwidth]{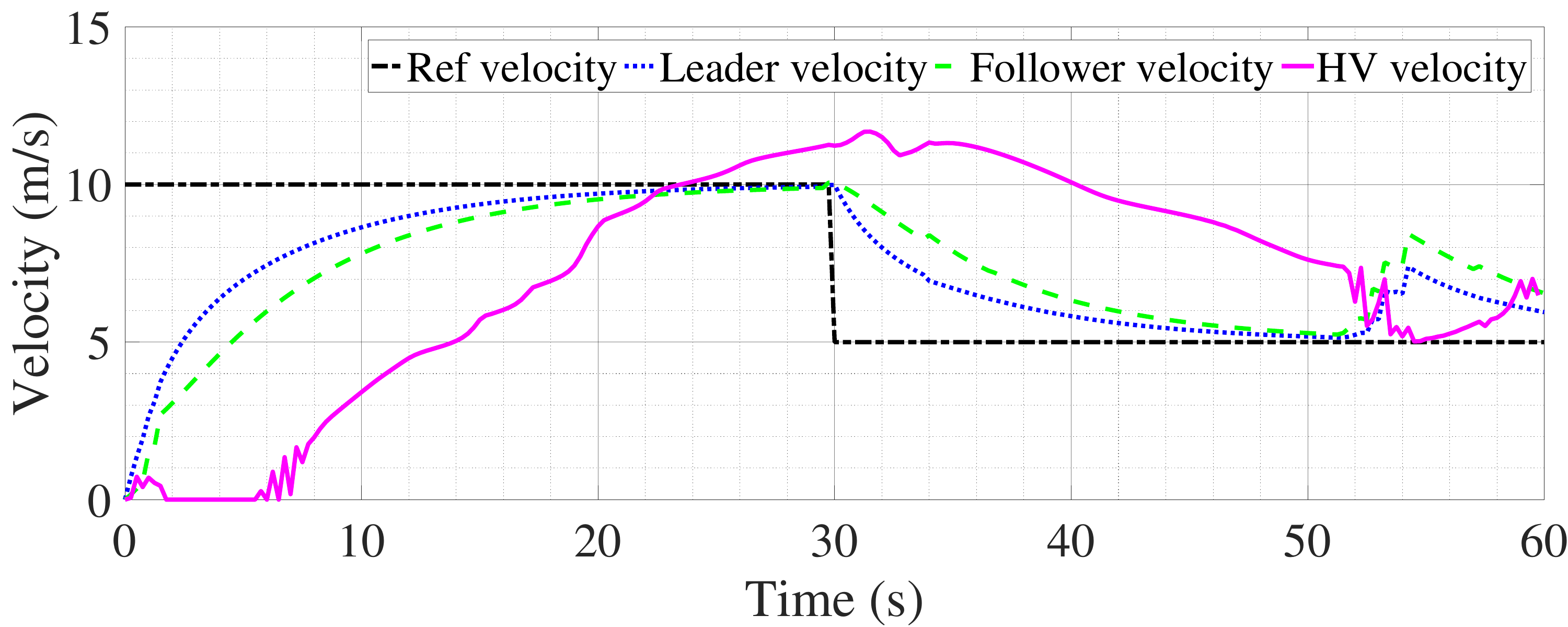} }}
    \qquad \qquad 
    \subfloat{{\includegraphics[trim=0cm 0cm 0cm 0.0cm, width=0.98\columnwidth]{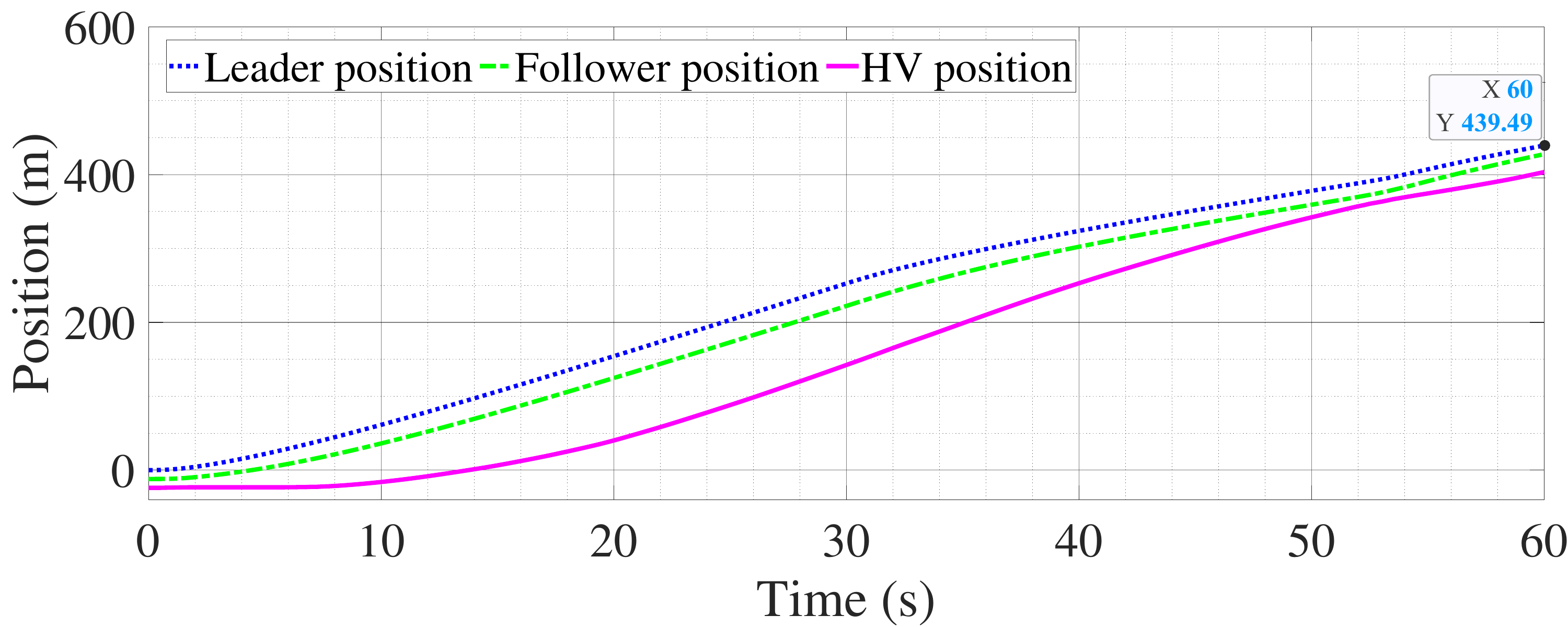} }}
    \qquad \qquad 
    \subfloat{{\includegraphics[trim=0cm 0cm 0cm 0.0cm, width=0.98\columnwidth]{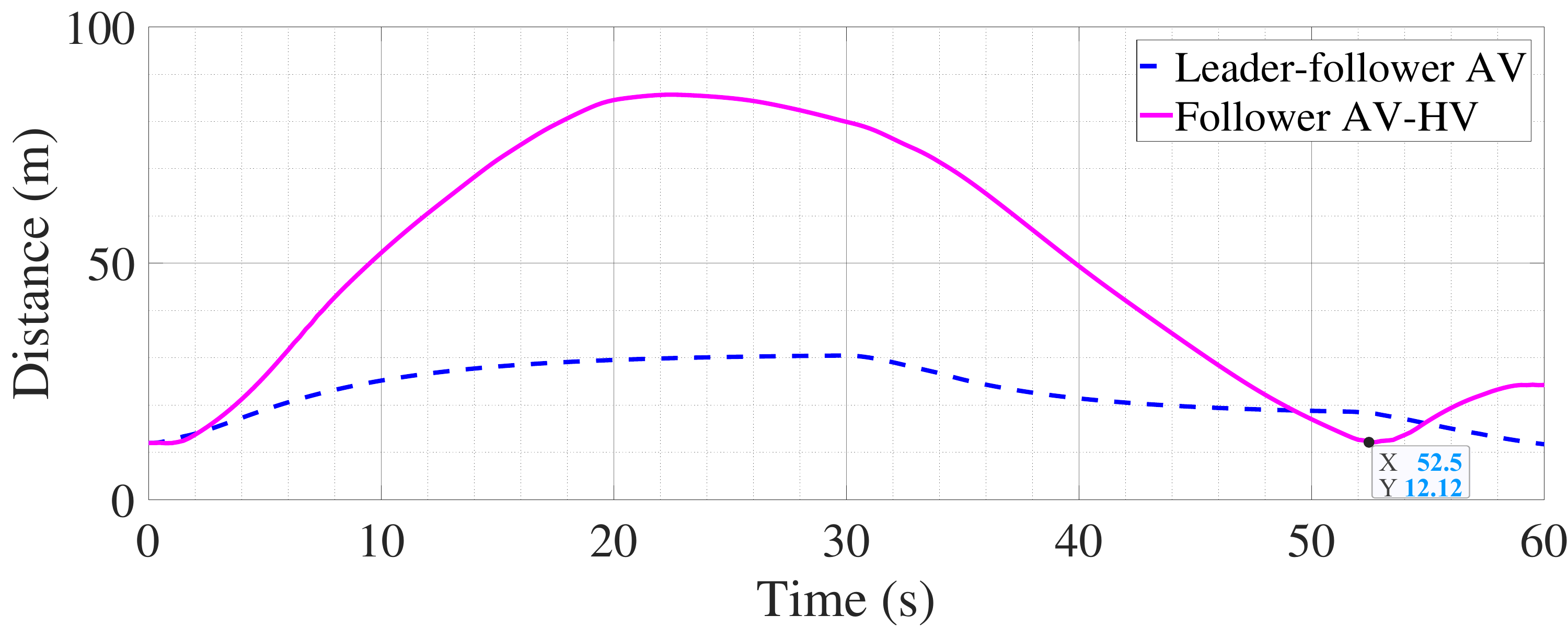} }}
    \caption{The simulation results of a low-speed emergency braking scenario using the GP-MPC in a real-time setting. These plots, sequentially represent velocity response, the position trajectories, and the inter-vehicle distance. Sequentially, the plots display velocity response, position trajectories, and inter-vehicle distances. Compared to the nominal MPC (Fig. \ref{figure:nominal_simulation_realTime}), the GP-MPC demonstrates improved performance by increasing the minimum distance between the HV and the trailing AV by 1\,m. This highlights the GP-MPC's ability to enhance safety while facilitating efficient vehicle movement in mixed-traffic scenarios.}
\label{figure:gp_simulation_realTime}
\end{figure}

In these real-time tests, we simulated a low-speed emergency braking scenario using both the nominal MPC and the GP-MPC. The leader AV's reference velocity was set to decrease from 10\,m$/$s to 5\,m$/$s at 30\,s. The results, as illustrated in Fig. \ref{figure:nominal_simulation_realTime} for the nominal MPC and Fig. \ref{figure:gp_simulation_realTime} for the GP-MPC, encompassed velocity response, position trajectories, and inter-vehicle distances. These figures demonstrate similar performance improvement with the GP-MPC compared to the nominal MPC as shown in Sec. \ref{sec:emergency} and \ref{sec:wltp}. The GP-MPC maintained a minimum HV-AV distance of 12.12\,m, which is approximately 1\,m greater than the nominal MPC. Additionally, vehicles under the GP-MPC consistently led those under the nominal MPC, indicating not only improved safety but also more efficient traffic management in the mixed platoon, particularly in this real-time operational setting. This demonstration of the GP-MPC's effectiveness in real-time scenarios, contributing to safer and more efficient traffic management, underscores its potential practical applicability in mixed-traffic environments.

Evaluating computational times is particularly essential in real-time implementation scenarios. We compared the computational times of both the GP-MPC and the nominal MPC across each time step, summarizing these findings in Tab. \ref{tab:simulation_time}. This table compares the average (Ave Time), maximum (Max Time), and standard deviation (Time Std) of computational times for both controllers. Notably, the GP-MPC's average computational time is only slightly longer (4.6\%) than the nominal MPC, a minor difference that maintains operational feasibility. In fact, the average computation time for the sparse GP-MPC is approximately 100-fold faster compared to our preliminary work \cite{wang2024improving} that did not implement these techniques. Its maximum computational time, 0.2245\,s, supports the potential for real-time operation at 4\,Hz. Moreover, the GP-MPC's more consistent computation time, as indicated by a slightly smaller standard deviation, highlights its stability compared to the nominal MPC. This comparison is crucial to assess the feasibility and efficiency of the GP-MPC in real-time settings and understand how it compares to the nominal MPC in terms of computational demand. 

\begin{table}
    \vspace{0.1cm}
    \caption{Computation time summary: The averaged time at each time step of the GP-MPC is only 4.6\% more than the nominal MPC. 
    }
    \label{tab:simulation_time}
    \begin{center}
        \begin{tabular}{|c|c|c|c|}
            \hline Controller & Ave Time & Max Time & Time Std \\
            \hline Nom. MPC & \textbf{0.1923}\,s & \textbf{0.2108}\,s & 0.0053 \\
            \hline GP-MPC & 0.2015\,s & 0.2245\,s & \textbf{0.0046} \\
            \hline
        \end{tabular}
    \end{center}
\end{table}

Despite these promising results, it is important to note limitations in high-speed scenarios. Tests revealed that performance decreases when the reference speed before emergency braking exceeds 10\,m$/$s. This is attributed to the larger sample time, which results in less frequent control updates. This can lead to diminished responsiveness to system changes or disturbances, making it difficult for the controller to accurately track the reference velocity in challenging dynamic environments like emergency braking. This real-time testing underscores the GP-MPC's potential applicability in actual hardware. However, further refinement is needed to enhance the controller's adaptability and responsiveness, especially in higher-speed situations. 

\subsection{Summary}
\label{sec:discussions_simulations}
This section has demonstrated the efficacy and practical applicability of the sparse Gaussian process-based model predictive control (GP-MPC) strategy in managing mixed-traffic platoons. Through our extensive simulations, including emergency braking scenarios, WLTP profiles, and real-time operational tests, we have evidenced significant improvements in safety, travel speed, and computational efficiency offered by the GP-MPC.

A primary insight from our simulations is the consistent superiority of the GP-MPC over the nominal MPC in maintaining safer HV-AV distances and facilitating more efficient vehicle movement within mixed platoons. In emergency braking scenarios, the GP-MPC enhanced the minimum HV-AV distance by approximately 0.9 meters, and in WLTP tests, it recorded an improvement of about 0.4 meters. These results highlight the proficiency of our GP-based HV model in dynamically adjusting safety margins in response to estimated HV uncertainties across various traffic conditions. Moreover, the GP-MPC strategy not only bolsters safety but also contributes to higher overall vehicle speeds in the platoon, leading to optimized traffic flow and efficiency. Furthermore, the real-time tests validate the potential of the GP-MPC for real-world applications. Although there is a need for ongoing refinements, particularly in enhancing the computational speed within the MPC loop, our work has effectively narrowed the gap between theoretical concepts and practical implementations.

One challenge was conducting direct quantitative comparisons between our GP-MPC method and other advanced MPC strategies, such as those discussed in Sec. \ref{sec:related_work}. These methods have distinct objectives and operate under varying assumptions, which complicates a straightforward comparison. Each strategy is tailored to specific aspects of mixed traffic control, highlighting the necessity for a nuanced understanding of the strengths and limitations inherent in each approach. This points towards the need for custom solutions in the diverse field of mixed-traffic control. Looking ahead, integrating the GP-MPC method with other advanced MPC strategies opens exciting avenues for future research. Such integrations could potentially combine the strengths of various approaches. For instance, combining the robustness of probabilistic bounds in Tube MPC \cite{feng2021robust} and the scalability of distributed MPC \cite{zhan2022data}, with the uncertainty management capabilities of our GP-MPC, could lead to more comprehensive and adaptive control strategies. This collaborative effort has the potential to significantly advance the safety and efficiency of mixed traffic systems, paving the way for more resilient and responsive traffic management solutions.

\section{Conclusion}
\label{sec:conclusion}

This article proposes an innovative approach to modeling human-driven vehicles (HVs) by combining a first-principles nominal model with a Gaussian process learning-based component. This model enhances not only the accuracy of the HV behavior prediction but also provides an estimate of HV uncertainty, which consequently improves the control of mixed vehicle platoons. A model predictive control policy has been developed based on this model, and its utility has been assessed in longitudinal car-following situations within mixed platoons. Compared to a standard MPC approach, extensive simulation studies conducted in this paper revealed that the proposed policy increases safety and enables a more efficient motion pattern for vehicles in mixed platoons. By incorporating a dynamic sparse GP technique in the MPC prediction loop, our MPC model effectively utilizes computational resources, requiring only a slight increase of 4.6\% in computational time in comparison to the standard MPC. This denotes a considerable decrease in computational time compared to our preceding work \cite{wang2024improving}, with a marked improvement of approximately 100 times. 

Our work has certain limitations that need to be addressed in further studies. Firstly, the current work focuses on longitudinal car-following scenarios where a human-driven vehicle follows an autonomous vehicle platoon. Extending the current research to include more complex traffic scenarios such as the inclusion of human-driven vehicles within the platoon or the incorporation of merging and lane-changing maneuvers would be a worthwhile endeavor. In addition, another key direction for future research involves expanding the scope of the controller design to encompass system utility. This involves developing a sophisticated multi-objective optimization algorithm within the GP-MPC framework that can balance individual vehicle safety and overall traffic efficiency. Such advancements could significantly contribute to the smart management of mixed traffic environments, paving the way for more sustainable and efficient urban mobility solutions. Furthermore, deeper research into the stability analysis of GP-based MPC could enhance the theoretical understanding and potentially broaden the application of GP-based MPC in real-world scenarios. Last but not least, even though custom solutions are needed in the diverse field of mixed-traffic control, exploring the integration of the GP-MPC method with other advanced MPC strategies could potentially further advance the safety and efficiency of mixed-traffic systems.

\section*{Acknowledgments}
This work was supported in part by the Natural Sciences and Engineering Research Council of Canada through the Discovery Grant program and by Magna International. We would like to express our sincere gratitude to Chris Tseng, Jiaming Zhong, Yukun Lu, Kunal Chandan, and Soroush Mortazavi for their exceptional contributions to our data collection process. Additionally, we extend our thanks to Dr. Zhihao Jiang with ShanghaiTech University, for providing valuable human-in-the-loop simulator data that significantly enriched our research.

\bibliographystyle{IEEEtran}
\bibliography{bibliography}

\begin{thebibliography}{10}
\providecommand{\url}[1]{#1}
\csname url@samestyle\endcsname
\providecommand{\newblock}{\relax}
\providecommand{\bibinfo}[2]{#2}
\providecommand{\BIBentrySTDinterwordspacing}{\spaceskip=0pt\relax}
\providecommand{\BIBentryALTinterwordstretchfactor}{4}
\providecommand{\BIBentryALTinterwordspacing}{\spaceskip=\fontdimen2\font plus
\BIBentryALTinterwordstretchfactor\fontdimen3\font minus
  \fontdimen4\font\relax}
\providecommand{\BIBforeignlanguage}[2]{{%
\expandafter\ifx\csname l@#1\endcsname\relax
\typeout{** WARNING: IEEEtran.bst: No hyphenation pattern has been}%
\typeout{** loaded for the language `#1'. Using the pattern for}%
\typeout{** the default language instead.}%
\else
\language=\csname l@#1\endcsname
\fi
#2}}
\providecommand{\BIBdecl}{\relax}
\BIBdecl

\bibitem{guo2023study}
X.-Y. Guo, G.~Zhang, and A.-F. Jia, ``Study on mixed traffic of autonomous
  vehicles and human-driven vehicles with different cyber interaction
  approaches,'' \emph{Vehicular Communications}, vol.~39, p. 100550, 2023.

\bibitem{du2021cooperative}
R.~Du, S.~Chen, Y.~Li, P.~Y.~J. Ha, J.~Dong, P.~C. Anastasopoulos, and S.~Labi,
  ``A cooperative crash avoidance framework for autonomous vehicle under
  collision-imminent situations in mixed traffic stream,'' in \emph{2021 IEEE
  International Intelligent Transportation Systems Conference (ITSC)}.\hskip
  1em plus 0.5em minus 0.4em\relax IEEE, 2021, pp. 1997--2002.

\bibitem{du2022framework}
R.~Du, S.~Chen, Y.~Li, M.~Alinizzi, and S.~Labi, ``A framework for lane-change
  maneuvers of connected autonomous vehicles in a mixed-traffic environment,''
  \emph{Electronics}, vol.~11, no.~9, p. 1350, 2022.

\bibitem{huang2020learning}
M.~Huang, Z.-P. Jiang, and K.~Ozbay, ``Learning-based adaptive optimal control
  for connected vehicles in mixed traffic: robustness to driver reaction
  time,'' \emph{IEEE transactions on cybernetics}, vol.~52, no.~6, pp.
  5267--5277, 2020.

\bibitem{petrovic2020}
P.~Dorde, M.~Radomir, and D.~Pesic, ``Traffic accidents with autonomous
  vehicles: type of collisions, manoeuvres and errors of conventional
  vehicles’ drivers,'' \emph{Transportation research procedia}, vol.~45, pp.
  161--168, 2020.

\bibitem{xue2023}
Y.~Xue, X.~Zhang, Z.~Cui, B.~Yu, and K.~Gao, ``A platoon-based cooperative
  optimal control for connected autonomous vehicles at highway on-ramps under
  heavy traffic,'' \emph{Transportation research part C: Emerging
  technologies}, vol. 150, p. 104083, 2023.

\bibitem{chen2021}
C.~Chen, J.~Wang, Q.~Xu, J.~Wang, and K.~Li, ``Mixed platoon control of
  automated and human-driven vehicles at a signalized intersection: dynamical
  analysis and optimal control,'' \emph{Transportation research part C:
  Emerging technologies}, vol. 127, p. 103138, 2021.

\bibitem{kessels2019traffic}
F.~Kessels, R.~Kessels, and Rauscher, \emph{Traffic flow modelling}.\hskip 1em
  plus 0.5em minus 0.4em\relax Springer, 2019.

\bibitem{li2023survey}
J.~Li, C.~Yu, Z.~Shen, Z.~Su, and W.~Ma, ``A survey on urban traffic control
  under mixed traffic environment with connected automated vehicles,''
  \emph{Transportation research part C: Emerging technologies}, vol. 154, p.
  104258, 2023.

\bibitem{guo2020}
L.~Guo and Y.~Jia, ``Inverse model predictive control {(IMPC)} based modeling
  and prediction of human-driven vehicles in mixed traffic,'' \emph{IEEE
  Transactions on Intelligent Vehicles}, vol.~6, no.~3, pp. 501--512, 2020.

\bibitem{di2021survey}
X.~Di and R.~Shi, ``A survey on autonomous vehicle control in the era of
  mixed-autonomy: From physics-based to ai-guided driving policy learning,''
  \emph{Transportation research part C: Emerging technologies}, vol. 125, p.
  103008, 2021.

\bibitem{morton2016}
J.~Morton, T.~A. Wheeler, and M.~J. Kochenderfer, ``Analysis of recurrent
  neural networks for probabilistic modeling of driver behavior,'' \emph{IEEE
  Transactions on Intelligent Transportation Systems}, vol.~18, no.~5, pp.
  1289--1298, 2016.

\bibitem{qu2017}
T.~Qu, S.~Yu, Z.~Shi, and H.~Chen, ``Modeling driver's car-following behavior
  based on hidden {Markov} model and model predictive control: A cyber-physical
  system approach,'' in \emph{2017 11th Asian Control Conference (ASCC)}.\hskip
  1em plus 0.5em minus 0.4em\relax IEEE, 2017, pp. 114--119.

\bibitem{lefevre2014a}
S.~Lefevre, Y.~Gao, D.~Vasquez, H.~E. Tseng, R.~Bajcsy, and F.~Borrelli, ``Lane
  keeping assistance with learning-based driver model and model predictive
  control,'' in \emph{12th International Symposium on Advanced Vehicle
  Control}, 2014.

\bibitem{hewing2019}
L.~Hewing, J.~Kabzan, and M.~N. Zeilinger, ``Cautious model predictive control
  using {Gaussian} process regression,'' \emph{IEEE Transactions on Control
  Systems Technology}, vol.~28, no.~6, pp. 2736--2743, 2019.

\bibitem{haninger2022}
K.~Haninger, C.~Hegeler, and L.~Peternel, ``Model predictive control with
  {Gaussian} processes for flexible multi-modal physical human robot
  interaction,'' in \emph{2022 International Conference on Robotics and
  Automation (ICRA)}.\hskip 1em plus 0.5em minus 0.4em\relax IEEE, 2022, pp.
  6948--6955.

\bibitem{wang2024improving}
J.~Wang, Z.~Jiang, and Y.~V. Pant, ``Improving safety in mixed traffic: A
  learning-based model predictive control for autonomous and human-driven
  vehicle platooning,'' \emph{Knowledge-Based Systems}, vol. 293, p. 111673,
  2024.

\bibitem{wang2023learning}
J.~Wang, M.~T. Fader, and J.~A. Marshall, ``Learning-based model predictive
  control for improved mobile robot path following using {Gaussian processes}
  and feedback linearization,'' \emph{Journal of Field Robotics}, vol.~40, pp.
  1014--1033, 2023.

\bibitem{guo2021anticipative}
L.~Guo and Y.~Jia, ``Anticipative and predictive control of automated vehicles
  in communication-constrained connected mixed traffic,'' \emph{IEEE
  Transactions on Intelligent Transportation Systems}, vol.~23, no.~7, pp.
  7206--7219, 2021.

\bibitem{feng2021robust}
S.~Feng, Z.~Song, Z.~Li, Y.~Zhang, and L.~Li, ``Robust platoon control in mixed
  traffic flow based on tube model predictive control,'' \emph{IEEE
  Transactions on Intelligent Vehicles}, vol.~6, no.~4, pp. 711--722, 2021.

\bibitem{zhan2022data}
J.~Zhan, Z.~Ma, and L.~Zhang, ``Data-driven modeling and distributed predictive
  control of mixed vehicle platoons,'' \emph{IEEE Transactions on Intelligent
  Vehicles}, vol.~8, no.~1, pp. 572--582, 2022.

\bibitem{snelson2005}
E.~Snelson and Z.~Ghahramani, ``Sparse gaussian processes using
  pseudo-inputs,'' \emph{Advances in neural information processing systems},
  vol.~18, 2005.

\bibitem{wang2024learning-based}
J.~Wang, Y.~V. Pant, and Z.~Jiang, ``Learning-based modeling of
  human-autonomous vehicle interaction for improved safety in mixed-vehicle
  platooning control,'' \emph{Transportation Research Part C: Emerging
  Technologies}, vol. 162, p. 104600, 2024.

\bibitem{pirani2022}
M.~Pirani, Y.~She, R.~Tang, Z.~Jiang, and Y.~V. Pant, ``Stable interaction of
  autonomous vehicle platoons with human-driven vehicles,'' in \emph{2022
  American Control Conference (ACC)}.\hskip 1em plus 0.5em minus 0.4em\relax
  IEEE, 2022, pp. 633--640.

\bibitem{macadam2003}
C.~C. {MacAdam}, ``{Understanding and Modeling the Human Driver},''
  \emph{Vehicle System Dynamics}, 2003.

\bibitem{wang2023intuitive}
J.~Wang, ``An intuitive tutorial to {Gaussian} process regression,''
  \emph{Computing in Science \& Engineering}, vol.~25, no.~4, pp. 4--11, 2023.

\bibitem{rasmussen2006}
C.~E. Rasmussen and C.~K.~I. Williams, \emph{{Gaussian} processes in machine
  learning}.\hskip 1em plus 0.5em minus 0.4em\relax MIT Press, 2006.

\bibitem{wiseadsmooseweb}
M.~Antkiewicz, ``{WISE} automated driving system and {UW Moose},''
  \url{https://uwaterloo.ca/waterloo-intelligent-systems-engineering-lab/projects/wise-automated-driving-system-and-uw-moos},
  2023.

\bibitem{pitropov2021}
M.~Pitropov, D.~E. Garcia, J.~Rebello, M.~Smart, C.~Wang, K.~Czarnecki, and
  S.~Waslander, ``Canadian adverse driving conditions dataset,'' \emph{The
  International Journal of Robotics Research}, vol.~40, no. 4-5, pp. 681--690,
  2021.

\bibitem{openpcdet2020}
O.~D. Team, ``{OpenPCDet}: An open-source toolbox for {3D} object detection
  from point clouds,'' \url{https://github.com/open-mmlab/OpenPCDet}, 2020.

\bibitem{antkiewicz2020}
\BIBentryALTinterwordspacing
M.~Antkiewicz, M.~Kahn, M.~Ala, K.~Czarnecki, P.~Wells, A.~Acharya, and
  S.~Beiker, ``Modes of automated driving system scenario testing: Experience
  report and recommendations,'' \emph{SAE Int. J. Adv. \& Curr. Prac. in
  Mobility}, vol.~2, pp. 2248--2266, 04/2020 2020. [Online]. Available:
  \url{https://saemobilus.sae.org/content/2020-01-1204}
\BIBentrySTDinterwordspacing

\bibitem{zhong2023hierarchical}
J.~Zhong, Z.~He, J.~Wang, and J.~Xie, ``A hierarchical framework for passenger
  inflow control in metro system with reinforcement learning,'' \emph{IEEE
  Transactions on Intelligent Transportation Systems}, vol.~24, no.~10, pp.
  10\,895--10\,911, 2023.

\bibitem{qadri2020state}
S.~S. S.~M. Qadri, M.~A. G{\"o}k{\c{c}}e, and E.~{\"O}ner, ``State-of-art
  review of traffic signal control methods: challenges and opportunities,''
  \emph{European transport research review}, vol.~12, pp. 1--23, 2020.

\bibitem{gueriau2020quantifying}
M.~Gu{\'e}riau and I.~Dusparic, ``Quantifying the impact of connected and
  autonomous vehicles on traffic efficiency and safety in mixed traffic,'' in
  \emph{2020 IEEE 23rd International Conference on Intelligent Transportation
  Systems (ITSC)}.\hskip 1em plus 0.5em minus 0.4em\relax IEEE, 2020, pp. 1--8.

\bibitem{dong2022impact}
J.~Dong, J.~Wang, and D.~Luo, ``Impact of connected and autonomous vehicles on
  traffic safety of mixed traffic flow: from the perspective of connectivity
  and spatial distribution,'' \emph{Transportation safety and environment},
  vol.~4, no.~3, p. tdac021, 2022.

\bibitem{berberich2020}
J.~Berberich, J.~K{\"o}hler, M.~A. M{\"u}ller, and F.~Allg{\"o}wer,
  ``Data-driven model predictive control with stability and robustness
  guarantees,'' \emph{IEEE Transactions on Automatic Control}, vol.~66, no.~4,
  pp. 1702--1717, 2020.

\bibitem{bujarbaruah2020}
M.~Bujarbaruah, X.~Zhang, M.~Tanaskovic, and F.~Borrelli, ``Adaptive stochastic
  {MPC} under time-varying uncertainty,'' \emph{IEEE Transactions on Automatic
  Control}, vol.~66, no.~6, pp. 2840--2845, 2020.

\bibitem{zhang2022}
T.~Zhang, S.~Li, and Y.~Zheng, ``Implementable stability guaranteed
  {Lyapunov}-based data-driven model predictive control with evolving gaussian
  process,'' \emph{Industrial \& Engineering Chemistry Research}, vol.~61,
  no.~39, pp. 14\,681--14\,690, 2022.

\bibitem{coppola2022eco}
A.~Coppola, D.~G. Lui, A.~Petrillo, and S.~Santini, ``Eco-driving control
  architecture for platoons of uncertain heterogeneous nonlinear connected
  autonomous electric vehicles,'' \emph{IEEE Transactions on Intelligent
  Transportation Systems}, vol.~23, no.~12, pp. 24\,220--24\,234, 2022.

\bibitem{zhu2023vehicle}
B.~Zhu, ``Vehicle performance simulation via {MATLAB},''
  https://github.com/tomzhu0225/vehicle-performance-simulation-via-matlab,
  2023.

\end{thebibliography}

\end{document}